\documentclass[10pt,twocolumn,letterpaper]{article}

\usepackage[pagenumbers]{cvpr} % To force page numbers, e.g. for an arXiv version

\DeclareMathOperator*{\diag}{diag}

\usepackage{multirow}

\definecolor{cvprblue}{rgb}{0.21,0.49,0.74}
\usepackage[pagebackref,breaklinks,colorlinks,allcolors=cvprblue]{hyperref}

\usepackage{algorithm}
\usepackage{algpseudocode}
\usepackage{amsmath}
\usepackage{cases}
\usepackage{bm}
\usepackage[frozencache,cachedir=minted-cache]{minted}
\usepackage[table]{xcolor}
\def\paperID{} % *** Enter the Paper ID here
\def\confName{CVPR}
\def\confYear{2026}

\title{VMonarch: Efficient Video Diffusion Transformers with Structured Attention}

\author{%
Cheng Liang$^{1,3}$\thanks{Work is done during internship at Kling Team, Kuaishou Technology.}, 
Haoxian Chen$^{1}$, 
Liang Hou$^{3}$, 
Qi Fan$^{2}$, 
Gangshan Wu$^{1}$, 
Xin Tao$^{3}$, 
Limin Wang$^{1}$\thanks{Corresponding author.}\\
$^{1}$State Key Laboratory for Novel Software Technology, Nanjing University   \quad \\
$^{2}$ School of Intelligence Science and Technology, Nanjing University  \quad \\
$^{3}$ Kling Team, Kuaishou Technology\\
{\tt\small \texttt{chienliang03@gmail.com} \texttt{lmwang@nju.edu.cn}
}
}

\begin{document}

\maketitle
\begin{abstract}
The quadratic complexity of the attention mechanism severely limits the context scalability of Video Diffusion Transformers (DiTs). We find that the highly sparse spatio-temporal attention patterns exhibited in Video DiTs can be naturally represented by the Monarch matrix. It is a class of structured matrices with flexible sparsity, enabling sub-quadratic attention via an alternating minimization algorithm. Accordingly, we propose VMonarch, a novel attention mechanism for Video DiTs that enables efficient computation over the dynamic sparse patterns with structured Monarch matrices. First, we adapt spatio-temporal Monarch factorization to explicitly capture the intra-frame and inter-frame correlations of the video data. Second, we introduce a recomputation strategy to mitigate artifacts arising from instabilities during alternating minimization of Monarch matrices. Third, we propose a novel online entropy algorithm fused into FlashAttention, enabling fast Monarch matrix updates for long sequences. Extensive experiments demonstrate that VMonarch achieves comparable or superior generation quality to full attention on VBench after minimal tuning. It overcomes the attention bottleneck in Video DiTs, reduces attention FLOPs by a factor of  $17.5$, and achieves a speedup of over $5\times$ in attention computation for long videos, surpassing state-of-the-art sparse attention methods at 90\% sparsity. Code is available at \url{https://github.com/EricLina/VMonarch}.
\vspace{-4mm} 
\end{abstract}
\vspace{-1mm}
\section{Introduction}
\vspace{-1mm}
\label{sec:intro}

\begin{figure*}[t]
    \vspace{-5mm} 
    \centering
    \includegraphics[width=0.9\linewidth]{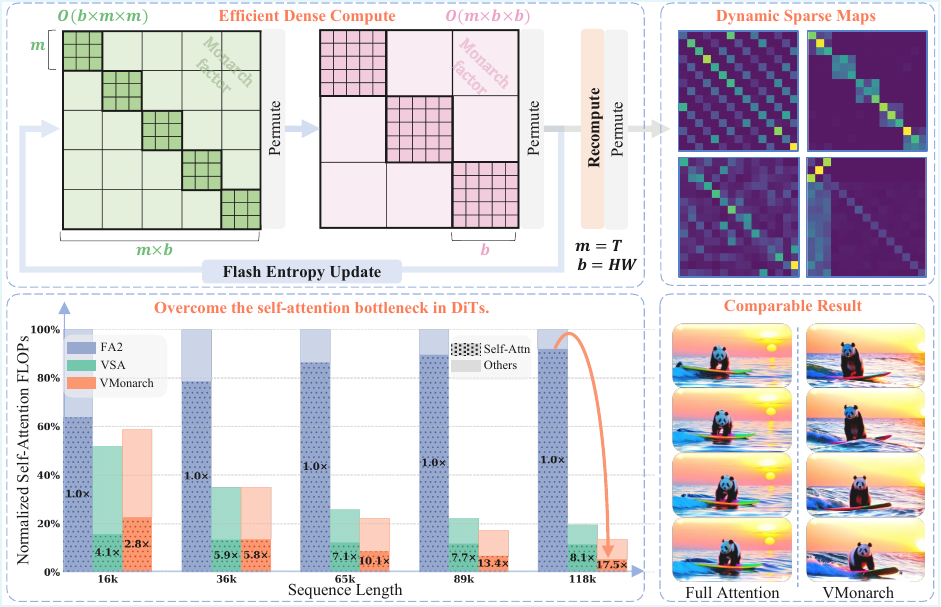}
    \caption{
    Video Monarch Attention (VMonarch) leverages dynamically updated spatial-temporal Monarch matrices~\cite{dao2022monarch} to model sparse attention patterns in video DiTs. 
    When generating a 321-frame video at 448$\times$832 resolution (approximately 118K tokens), VMonarch speeds up self-attention computation by 17.5$\times$ compared to full attention (FA2)~\cite{dao2023flashattention} and by 2$\times$ compared to highly sparse alternatives like Video Sparse Attention (VSA) at 90\% sparsity.
    After minimal fine-tuning, VMonarch attains comparable video generation quality to FA2.
    }
    \label{fig:teaser}
    \vspace{-1mm} 
\end{figure*}

% What is the problem
%%%-------Part 1: Video Generation's bottleneck in self attention computation -------%%%
Long context modeling is a critical challenge in modern deep learning, particularly in Video Diffusion Transformers (DiTs) aiming to generate long-duration videos~\cite{wan2025wan,kong2024hunyuanvideo,yang2024cogvideox,ma2024latte,yin2025slow}. Video DiTs face significant computational bottlenecks due to the quadratic complexity of vanilla attention. According to Wan‑2.1~\cite{wan2025wan}, the attention mechanism accounts for up to 95\% of the total computation at a sequence length of one million tokens.

% 1.1 sub-quadratic attention Mechanisms
To address the efficiency issue of attention mechanisms in modeling long sequences, 
various sub-quadratic attention mechanisms~\cite{yuan2025native,xu2025xattention,yang2024gated,xi2025sparse,zhangspargeattention,lu2025moba,chen2025sana,choromanski2020rethinking,wang2020linformer,xiong2021nystromformer,dong2024flex} have been proposed.
They mitigate the $\mathcal{O}(N^2)$ complexity of vanilla attention, where $N$ represents the sequence length, and can be broadly categorized into two main types: Sparse Attention~\cite{yuan2025native,xu2025xattention} and Linear Attention~\cite{yang2024gated,chen2025sana}.
Sparse Attention mechanisms reduce computation by focusing on a subset of relevant tokens using predefined~\cite{xi2025sparse, sun2025vorta,chen2025sparse, dao2021pixelated} or dynamic blockwise sparse patterns~\cite{xu2025xattention,zhangspargeattention,zhang2025vsa,lu2025moba,wu2025vmoba}. The complexity of Sparse Attention is reduced to $\mathcal{O}(\tau N^{2})$, where $1-\tau$ is the sparsity ratio. However, improperly defined patterns or too aggressive sparsity ratios can lead to significant performance degradation~\cite{wu2025vmoba,zhang2025vsa}, 
and the actual efficiency gain is often less than expected due to overhead from managing sparse structures~\cite{wu2025vmoba} and irregular memory access~\cite{yang2025sparse}, which poses more challenges for hardware-efficient implementation~\cite{yuan2025native,dong2024flex}.
Linear Attention mechanisms~\cite{choromanski2020rethinking,wang2020linformer,xiong2021nystromformer} approximate the attention computation using kernel methods, reducing the complexity to $\mathcal{O}(N)$. While they offer significant efficiency improvements, Linear Attention methods often suffer from a performance gap compared to standard attention due to their low-rank structure, which limits their expressive power
~\cite{fan2025breaking,zhang2025sla}.

Sparse Attention and Linear Attention can be viewed as two matrix compression strategies that exploit sparsity or low-rank structures, respectively. Notably, the attention matrices in video DiTs are high-rank and sparse, exhibiting strong block-diagonal structures due to the inherent spatiotemporal locality in video data. This indicates that sparsity-based methods are more suitable for Video DiTs~\cite{xi2025sparse,wan2025wan}.
Unlike prior sparse attention methods that use ineffective fixed patterns or unstructured dynamic patterns that are inefficient to compute densely, MonarchAttention~\cite{yaras2025monarchattention} constructs the sparse attention maps using dynamically updated structured Monarch matrices~\cite{dao2022monarch}.
Monarch matrices are parameterized as the product of hardware-friendly block-diagonal matrices interleaved with permutations, and can model a wide class of linear transforms, such as Toeplitz matrices~\cite{gray2006toeplitz} and Butterfly matrices~\cite{dao2020kaleidoscope}.
By alternately updating two dual Monarch matrices of size $\sqrt{N}$ with an entropy-based regularization algorithm, Monarch Attention reduces the computational complexity to an optimal $\mathcal{O}(N\sqrt{N})$. This approach achieves a favorable balance between efficiency and expressiveness, making it particularly effective for modeling sparse structures.

% Challenges of applying Monarch attention to video DiTs
However, directly applying Monarch Attention to Video DiTs presents several challenges:
% 1.Split sequence into blocks along spatiotemporal dimensions
First, designing an effective block structure for Monarch matrices that captures both spatial and temporal dependencies in video is non-trivial, as the most efficient structure with monarch matrix sizes of $\sqrt{N}$ can disrupt the inherent spatiotemporal structure of video data.
% 2. First frame quality degradation in post-training
Second, the inherent attention sink phenomenon in Video DiTs~\cite{xi2025sparse} causes the entropy-based regularization in MonarchAttention to be disproportionately focused on the first frame. This results in an excessively large softmax temperature, thereby degrading generation quality.
% 3 Customized Kernel Implementation when handling video data
Third, existing Monarch Attention implementations are not optimized for long sequences. Large monarch matrix sizes result in significant memory overhead during the alternating maximization updates, hindering the practical application to video modalities with very long sequences.

In this work, we propose Video Monarch Attention (VMonarch), a novel and efficient framework that effectively integrates Monarch Attention into Video DiTs to address the aforementioned challenges. As shown in \cref{fig:teaser}, our key contributions are summarized as follows:
\begin{itemize}
    \item We are the first to explore representing the sparse attention maps in Video DiTs using Monarch matrices, for which we design a novel spatiotemporal block structure to capture dependencies in video data effectively, providing a new perspective on efficient attention mechanisms via structured matrix representations.
    \item To mitigate the quality degradation of the first frame due to excessive entropy regularization when updating Monarch matrices, we introduce a first-frame recomputation strategy that preserves the quality of the initial frame with modest computational overhead.
    \item We propose a novel online-entropy algorithm, analogous to the online softmax in FlashAttention~\cite{dao2023flashattention}, implemented in a customized GPU kernel to enable efficient Monarch matrix updating for long video sequences.
    \item Extensive experiments show that VideoMonarch attains comparable or superior generation quality to full attention on VBench after minimal fine-tuning, while achieving over $17.5 \times$ attention FLOPs reduction with more than $5 \times$ kernel speedup compared to FlashAttention-2~\cite{dao2023flashattention}.
\end{itemize}

\section{Related Work}
\label{sec:relatedwork}

\paragraph{Diffusion Model Inference Acceleration.}
% 1. Denoise step: noise schedules, distillation, cache
Accelerating diffusion model inference has been an active area of research. Some works focus on reducing the computational cost of the denoising steps required during inference
through techniques such as improved noise schedules~\cite{song2020denoising,ho2020denoising,lu2022dpm,liu2022flow,song2023consistency,luo2023latent}, knowledge distillation for fewer steps~\cite{yin2024improved,salimans2022progressive}, and reusing cached intermediate results~\cite{liu2025timestep,ma2024deepcache,lv2024fastercache}. Complementary to these methods, optimizing the attention mechanism with sparsification~\cite{xu2025xattention,xi2025sparse,zhang2025vsa,chen2025sparse,li2025radial,wu2025vmoba} or linearization~\cite{xie2024sana,zhang2025sla,hui2025arflow,ghafoorian2025attention} is an orthogonal approach that is crucial for accelerating diffusion model inference on long-sequence modalities, such as video, where attention computation becomes a significant bottleneck as the sequence length increases~\cite{wan2025wan} .

% 2. Attention acceleration (Current Bottleneck)
% 3. Quantized attention
\paragraph{Efficient Attention Mechanisms.}

Sparse Attention mechanisms reduce the quadratic complexity of standard self-attention by limiting the number of key-value pairs to which each query attends. These mechanisms can be broadly categorized based on their sparsity patterns. 
Some methods employ fixed patterns to reduce attention computation, including local windowed attention~\cite{beltagy2020longformer}, strided or dilated attention~\cite{zaheer2020big,beltagy2020longformer},  $\Lambda$-shape Pattern~\cite{xiao2024duoattention}, Vertical-Slash Pattern~\cite{jiang2024minference}, and so on. However, these fixed patterns may lack adaptability to varying input data.
Other methods utilize dynamic patterns that adapt to the input data, such as dynamic block sparse attention~\cite{xu2025xattention,dong2024flex,lai2025flexprefill,yuan2025native,lu2025moba,wu2025vmoba,sun2025vorta,zhang2025vsa,zhangspargeattention,shi2025trainable} and clustering-based attention mechanisms that regroup key-value pairs using k-means or locality-sensitive hashing (LSH)~\cite{chen2410magicpig,kitaev2020reformer,liu2025clusterkv}. 
% Some dynamic sparse attention methods combine both query (Q) and key-value (KV) sparsity for further efficiency gains~\cite{zhang2025training, zhan2025bidirectional}. While Q-sparsity is effective as a modality connector in multi-modal understanding tasks~\cite{li2023blip}, this downsampling of queries can lead to information loss, which is detrimental for high-fidelity generation tasks.
These dynamic methods often achieve better performance by tailoring the attention to the input, but they may incur additional computational overhead due to their unstructured dynamic patterns~\cite{yang2025sparse}.
% For example, SVG2~\cite{yang2025sparse} uses permutation operations to convert unstructured sparse attention patterns into a more contiguous format, enabling more efficient computation.
%
Another line of work focuses on Linear Attention mechanisms~\cite{xie2024sana,lieber2024jamba,li2025minimax,blakeman2025nemotron, huang2025m4v,fan2025breaking,yang2024gated,peng2023rwkv,yang2024gated}, which approximate the standard attention mechanism by reordering computations to achieve linear complexity with respect to the sequence length. Linear Attention demonstrates considerable inference efficiency (KV cache-free)~\cite{peng2023rwkv,yang2024gated} under causal settings and offers superior computation savings in non-causal settings~\cite{fan2025breaking,fan2025rectifying,han2024bridging,wei2025vit} due to its linear complexity. SANA-Video~\cite{chen2025sana} successfully applies blockwise linear attention to Video DiTs with a high-compression VAE. However, these methods are often limited by expressiveness due to their low-rank properties~\cite{zhang2025sla,fan2025breaking}, which need to be compensated for by hybrid architectures~\cite{lieber2024jamba,li2025minimax,blakeman2025nemotron, huang2025m4v}, rank augmentation techniques~\cite{fan2025breaking}, and state expansion~\cite{yang2024gated}.
Some works combine both sparse and linear attention mechanisms to model the high-rank and low-rank components of attention matrices, respectively~\cite{chen2021scatterbrain,zhang2025sla}. These hybrid approaches enable sparse attention to leverage a higher sparsity ratio by using linear attention to capture the global context, indicating the importance of leveraging the inherent structure within attention matrices.
%
% 1.4.1 Quantized Attention
There are also works exploring quantized attention mechanisms~\cite{zhang2025sageattention3,zhangsageattention2}, which reduce the precision of attention computation to lower bit-widths, thereby decreasing memory usage and computational load while maintaining performance. However, quantized methods are constrained by hardware precision formats and often suffer from accuracy degradation.

\paragraph{Monarch Matrix.}
Monarch matrices are a family of structured matrices that are hardware-efficient and expressive, as they are parameterized as the products of block-diagonal matrices interleaved with permutation operations. They can represent a wide class of transforms, including convolution, Hadamard transforms, Toeplitz matrices~\cite{gray2006toeplitz}, Butterfly matrices~\cite{dao2020kaleidoscope} and AFDF matrices~\cite{moczulski2015acdc}.
Previous works~\cite{dao2022monarch, fu2023monarch} have shown the efficiency and effectiveness of Monarch matrices by applying them to neural network sparsification.
Monarch matrices can be computed densely~\cite{fu2023monarch} by leveraging their block-diagonal structure, and they have a complexity from $\mathcal{O}(N \log N)$ to $\mathcal{O}(N^{3/2})$ depending on the number of Monarch factors~\cite{dao2022monarch}, offering flexible sparsity.
Moreover, MonarchAttention~\cite{yaras2025monarchattention} uses Monarch matrices to approximate the attention map using an alternating minimization method, which demonstrates effectiveness in modeling a sparse attention map.

\section{Method}
\label{sec:method}

\subsection{Preliminaries}
% Monarch Attention basics
\noindent\textbf{Monarch Matrix}
is a row-permuted block rank-one matrix~\cite{dao2022monarch}. Formally, given $N=m\times b$ for integers $m$ and $b$, we can define a Monarch matrix $\bm{M}\in\mathbb R^{N\times N}$ by
$$
\bm{M}=\bm{P}_{(b,N)}\bm{L}\bm{P}^\top_{(b,N)}\bm{R},
$$
where $\bm{L}$ and $\bm{R}$ are two block-diagonal matrices called Monarch factors: $\bm{L}=\text{diag}(\bm{L}_0,\dots,\bm{L}_{b-1})$ and $\bm{R}=\text{diag}(\bm{R}_0,\dots,\bm{R}_{m-1})$; $\bm{P}_{(b,N)}$ is a permutation matrix, whose ($i+1$)-th row is defined by $\bm{e}_{\sigma_{(b,N)}(i)}$ and $\sigma_{(b,N)}(i)=i\bmod b + \lfloor i/b\rfloor\cdot b$\footnote{Applying $\bm P_{(b, N)}$ to a length n vector means reshaping it into an $m\times b$ matrix in row-major order, transposing it, and then flattening it back to a vector in row-major order.}. Write $\bm{M}$ in block matrix form. ${M}_{ijkl}$ is the $(j,l)$ element in the $(i,k)$ matrix. Denote the $(j,k)$ element in $\bm{L}_i$ as ${L}_{ijk}$. It can be derived that ${M}_{jikl}={L}_{ijk}{R}_{kil}$.

\textbf{Monarch Attention}~\cite{yaras2025monarchattention}
uses the Monarch matrix to approximate attention in the Transformer~\cite{vaswani2017attention}, which means finding a Monarch matrix $\bm{M}\approx \text{softmax}(\bm{Q}\bm{K}^\top)$ and computing $\bm{O}=\bm{M}\bm{V}$ to obtain the approximate result. To avoid the computation of $\sigma(\bm{Q}\bm{K}^\top)$, MonarchAttention first rewrites softmax as an optimization problem.
\begin{equation}
    \sigma( \bm{Q}  \bm{K}^\top) = \underset{\bm{A} \in \Delta^{N \times N}}{\text{argmax}}\; \langle  \bm{A},  \bm{Q}  \bm{K}^\top \rangle + H(\bm{A}),
\end{equation}
where $\Delta^{N \times N}$ denotes the set of row-stochastic matrices, and $H(\bm{A})=-\sum_{ij}\bm{A}_{ij}\log \bm{A}_{ij}$ is Shannon entropy. When $\bm{A}$ is a Monarch matrix, we have 
\begin{small}
\begin{equation}
\begin{aligned}
\langle \bm{M},\bm{QK}^\top\rangle + H(&\bm{M})
=\langle \bm{PM},\bm{PQK}^\top\rangle + H(\bm{PM})\\
=&\langle \bm{PM},\bm{\tilde Q}\bm{K}^T\rangle + H(\bm{PM})\\
=&\sum\limits_{ik}\langle \bm{L}_{ik}\bm{R}_{ki}^\top, \bm{\tilde Q}_i\bm{K}_k^\top\rangle+H(\bm{L}_{ik}\bm{R}_{ki}^\top),
\end{aligned}
\end{equation}
\end{small}

\noindent where $\bm Q=\{\bm Q_i\}_{i=1}^b$ and $\bm K=\{\bm K_i\}_{i=1}^m$.
MonarchAttention uses alternating maximization to optimize the objective under the fact that when one component is fixed, the problem is concave. We can derive the closed-form solution.
{
\fontsize{9pt}{10pt}\selectfont
\begin{numcases}{}
\boldsymbol{R}=\text{softmax}_l(\sum\limits_v\bm\alpha_{\bm{R},kiv}\bm{K}_{klv}/\bm{c}_{\bm{R},ki}) \label{eq:R} \\ 
\bm\alpha_{\bm{R},kiv}=\sum\limits_j \boldsymbol{L}_{ikj}\bm{Q}_{ijv}, \ \ \ \ 
\bm{c}_{\bm{R},ki}=\sum\limits_j \boldsymbol{L}_{ikj}, \label{eq:ar}
\end{numcases}
}

{
\fontsize{9pt}{10pt}\selectfont
% \begin{small}
% \begin{equation} 
\begin{numcases}{}
\bm{L}=\text{softmax}_j(\sum\limits_v\bm\alpha_{\bm{L},ikv}\bm{Q}_{ijv}-\bm{c}_{\bm{L},ik}) \label{eq:L} \\
\bm\alpha_{\bm{L},ikv}=\sum\limits_l \boldsymbol{R}_{kil}\bm{K}_{klv}, \
\bm{c}_{\bm L,ik}=\sum\limits_l \boldsymbol{R}_{kil}\log\boldsymbol{R}_{kil}, \label{eq:al}
\end{numcases}
% \end{equation}
% \end{small}
}

where $\operatorname{softmax}_i$ means applying the softmax in dimension $i$, and $\bm Q_{ijv}$ refers to the $v$-th element in the $j$-th column of $\bm Q_i$. After $t$ iterations, the approximating result $\bm M^{(t)}$ is obtained using $\bm L^{(t)}, \bm R^{(t)}$, and the output of attention can be calculated by $\bm O=\bm{L}^{(t)}\bm{R}^{(t)}\bm V$. The complexity of calculating $\bm{R}$ and $\bm L$ is $\mathcal{O}(m(bmd))$ and $\mathcal{O}(b(mbd))$, respectively; therefore, the total complexity is $\mathcal{O}(tN(m+b)d)$. MonarchAttention can reduce the computational complexity from $\mathcal{O}(N^2 d)$ to $\mathcal{O}(tN(m+b)d)$, where $t$ is a small constant.

\begin{figure}[t]
    \centering
    \includegraphics[width=\linewidth]{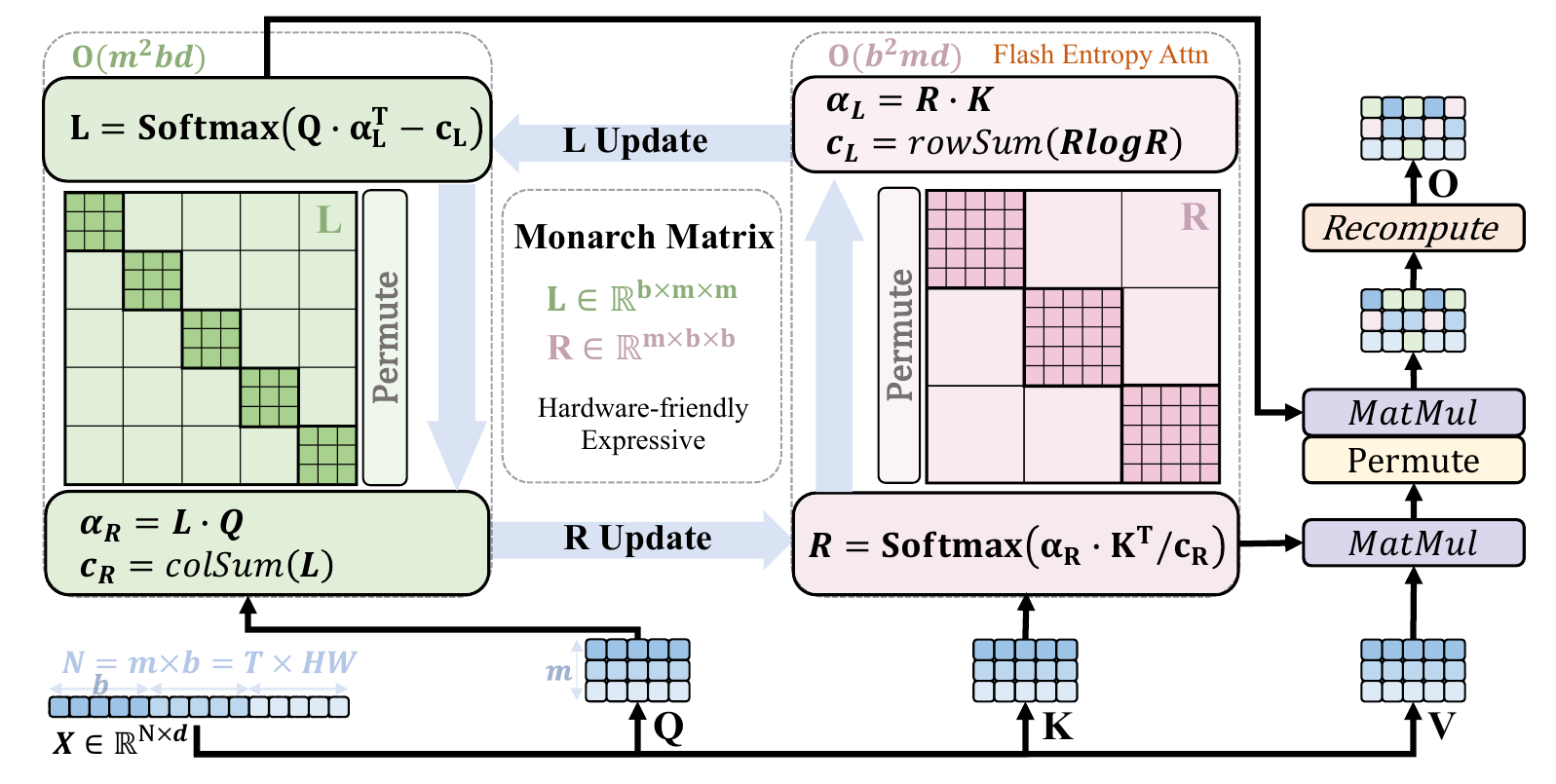}
    \caption{Overview of our Video Monarch Attention. (a) We represent the $N \times N$ full attention matrix via two alternatingly optimized smaller Monarch factors $\bm L$ and $\bm R$ with spatial-temporal structure factorization. (b) 
    We introduce a Recomputation Strategy to address quality degradation caused by excessively large temperature term $\bm{c}_{\bm{R}}$ for the first frame. 
    (c) We propose an Online-Entropy algorithm integrated with FlashAttention to accelerate the iterative computation of Monarch matrices further.}
    \label{fig:method_overview}
\end{figure}

\textbf{Implementation.}
As shown in~\cref{fig:method_overview}, the input matrices $\bm Q,\bm K,\bm V$ are first reshaped from $\mathbb R^{N\times d}$ to $\mathbb R^{m\times b\times d}$. The matrices $\bm Q$ and $\bm K$ are then used to calculate the monarch factors $\bm L$ and $\bm R$ by iteratively applying the closed-form solution. 
To minimize data movement and memory usage on GPUs, MonarchAttention combines (\cref{eq:R},~\cref{eq:al}) into a kernel and (\cref{eq:ar},~\cref{eq:L}) into another. Specifically, it stores $\bm\alpha_{\bm{L}}$, $\bm c_{\bm{L}}$, $\bm\alpha_{\bm{R}}$,$\bm c_{\bm{R}}$ in the loop and only writes $\bm L$ and $\bm R$ in the last iteration. After the iteration, the output of MonarchAttention can be calculated by the multiplication of $\bm L$, $\bm R$, and $\bm V$, interleaved with permutation.
The detailed Python-like code can be found in ~\cref{sec:code}.

\subsection{Video Monarch Attention}
% Video DiT has a Diagonal Attention map along spatio-temporal dimensions.
Video DiTs often exhibit a sparse, block-diagonal attention structure. Tokens within the same frame or adjacent pixels tend to have stronger interactions; this inherent structure aligns well with the sparse block-diagonal properties of Monarch matrices. To leverage this prior, we propose Video Monarch Attention, which adapts the Monarch matrix factorization with a spatio-temporal structure tailored for video DiTs.
For a video sequence consisting of $T$ frames, with each frame containing $H \times W$ spatial tokens, thus resulting in a total of $N=T H W$ tokens, we set the Monarch matrix parameters $m=T$ and $b=H \times W$. This aligns the block structure of the Monarch matrix with the spatio-temporal layout of video tokens.

Formally, given the queries $\bm{Q} \in \mathbb{R}^{N \times d}$, keys $\bm{K} \in \mathbb{R}^{N \times d}$, and values $\bm{V} \in \mathbb{R}^{N \times d}$, we compute the Video Monarch Attention output as:
\begin{equation}
    \label{eq:MonarchAttention_with_mbt}
    \bm{O} = \text{MonarchAttention}(\bm{Q}, \bm{K}, \bm{V}; m, b, t),
\end{equation}
where the MonarchAttention function utilizes the Monarch matrices defined by $m=T$ and $b=HW$ and performs $t=2$ iterations of alternating optimization.
% to efficiently compute the attention output $\bm{O} \in \mathbb{R}^{N \times d}$.
During the iterative optimization of the Monarch factors $\bm{L}$ and $\bm{R}$, we maintain the spatio-temporal structure by ensuring that $\bm{L} \in\mathbb{R}^{m^2 \times b}$ captures temporal dependencies across frames, while $\bm{R}\in\mathbb{R}^{b^2 \times m}$ captures spatial dependencies within each frame. This allows Video Monarch Attention to efficiently approximate the full attention map by factorizing it into dedicated spatial and temporal components while maintaining computational efficiency. 

\paragraph{Complexity Analysis.}
As we set $m=T$ and $b=H\times W$ to align with the spatio-temporal structure of video data, and utilize $t$ iterations for the alternating optimization, the computational complexity of Video Monarch Attention becomes $\mathcal{O}(tN(T + HW)d)$. Given $N=THW$ and $T \ll HW$, this yields a theoretical speedup of $\frac{THW}{t(T+HW)} \approx \frac{T}{t}$ over the standard $\mathcal{O}(N^2d)$ attention.

\subsection{First-Frame Recomputation Strategy}

\begin{figure}[t]
    \centering
    \includegraphics[width=1.0\linewidth]{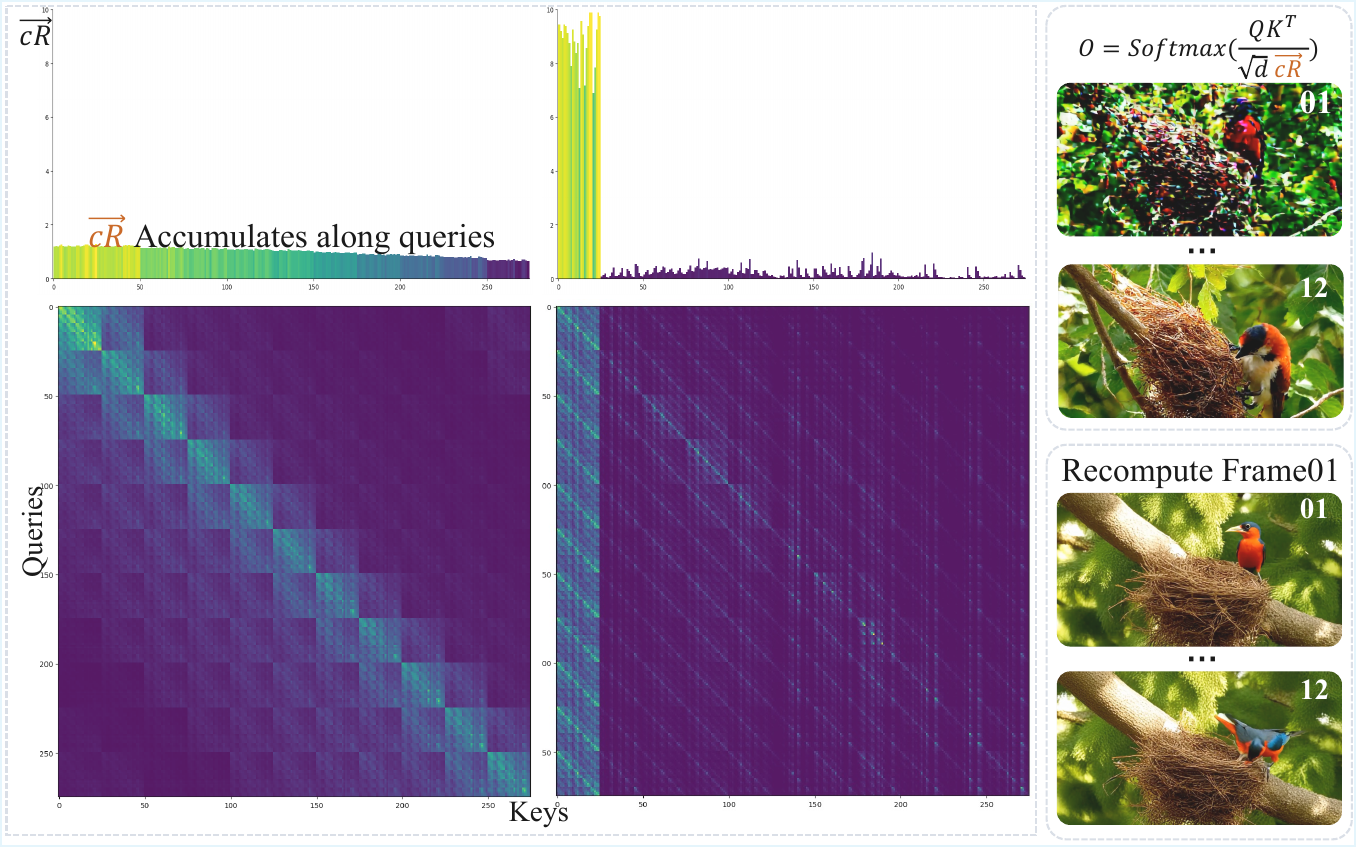}
    \caption{Illustration of the attention sink phenomenon in video models. The first frame tends to accumulate excessive attention from subsequent frames, leading to a loss of fine-grained details when using Monarch Attention. Our First-Frame Recomputation Strategy effectively restores these details by recalculating the attention for the first frame using full attention.}
    % W/ Sink: L28, Heads 9; 
    % W/O Sink: L28 Heads 10.
    \label{fig:first_frame_sink}
    \vspace{-1em}
\end{figure}

In video DiTs, there exists a well-known attention sink phenomenon in the first frame~\cite{xiao2023efficient,xi2025sparse,li2025radial}, where the initial frame serves as a crucial contextual anchor for the entire sequence and tends to attract disproportionate attention from subsequent frames. We observe that this attention sink phenomenon negatively impacts the performance of Video Monarch Attention, as shown in ~\cref{fig:first_frame_sink}. As the first frame tokens receive excessive cumulative attention score during the iterative optimization of Monarch Attention, the temperature adjustment term $\bm{c}_{\bm{R}}$ for these tokens becomes significantly large. This leads to a smoother attention distribution, causing degradation in the generation quality of the first frame.

To mitigate this issue, we introduce a simple yet effective First-Frame Recomputation Strategy. 
We explicitly recompute the attention outputs for the first frame using full attention, as shown in ~\cref{fig:method_overview}.
Specifically, we take the queries $\bm{Q}_0$ corresponding to the first frame, keys $\bm{K}$, and values $\bm{V}$, and compute the attention output as:
\begin{equation}
    \bm{O}_0 = \text{softmax}\left(\frac{\bm{Q}_0 \bm{K}^\top}{\sqrt{d}}\right) \bm{V}.
\end{equation}
This recomputation step introduces a modest computational overhead. The cost is $\mathcal{O}(bNd)$, specifically $\frac{b}{t(m+b)}$ of the total cost of Video Monarch Attention. 
This ensures that we can restore the fidelity of the first frame without compromising the overall efficiency of our method.

\subsection{Online Entropy Based FlashAttention}
\begin{algorithm}[!t]
\small 
\caption{Flash-{\color{blue}{Entropy}}-Attention}
\label{alg:flash_attention_entropy}
\begin{algorithmic}[1]
\Require $\mathbf{Q}, \mathbf{K}, \mathbf{V} \in \mathbb{R}^{N \times d}$, Block sizes $B_c, B_r$.
\Ensure $\mathbf{O} \in \mathbb{R}^{N \times d}$, $L \in \mathbb{R}^{N}$, {\color{blue}{${H} \in \mathbb{R}^{N}$ (Entropy)}}.
\State $T_r \gets \lceil N/B_r \rceil$, $T_c \gets \lceil N/B_c \rceil$.
\For{$i = 1 \to T_r$}
  \State Load $\mathbf{Q}_i$ to SRAM.
  \For{$j = 1 \to T_c$}
    \State Load $\mathbf{K}_j, \mathbf{V}_j$ to SRAM.
    \State $\mathbf{S}_{ij} \gets \mathbf{Q}_i \mathbf{K}_j^T$.
    \State $m_{i}^{(j)} \gets \max(m_{i}^{(j-1)}, \mathrm{rowmax}(\mathbf{S}_{ij}))$.
    \State $\mathbf{P}_{ij} \gets \exp(\mathbf{S}_{ij} - m_{i}^{(j)})$.
    \State $\alpha \gets \exp(m_{i}^{(j-1)} - m_{i}^{(j)})$.
    \State $\mathbf{O}_{i}^{(j)} \gets \diag(\alpha) \mathbf{O}_{i}^{(j-1)} + \mathbf{P}_{ij} \mathbf{V}_j$.
    \State $\ell_{i}^{(j)} \gets \alpha \odot \ell_{i}^{(j-1)} + \mathrm{rowsum}(\mathbf{P}_{ij})$.
    \State {\color{blue}{$h_i^{(j)} \gets \alpha \odot h_i^{(j-1)} + \alpha \log(\alpha) \odot \ell_i^{(j-1)}  + \mathrm{rowsum}(\mathbf{P}_{ij} \log(\mathbf{P}_{ij}))$}}.
  \EndFor
  \State $\mathbf{O}_{i} \gets \diag(\ell_{i}^{(T_c)})^{-1} \mathbf{O}_{i}^{(T_c)}$.
  \State $L_{i} \gets m_{i}^{(T_c)} + \log(\ell_i^{(T_c)})$.
  \State {\color{blue}{${H}_i \gets \log(\ell_i^{(T_c)}) - h_i^{(T_c)} / \ell_i^{(T_c)}$}}. 
  \State Write $\mathbf{O}_{i}, L_i, $ {\color{blue}{${H}_i$}} to HBM.
\EndFor
\end{algorithmic}
\end{algorithm}

Upon profiling Video Monarch Attention, we identify the primary computational bottleneck in the update step for the $\bm{R}$ matrix and its entropy term $\bm c_L$ (~\cref{eq:R}, ~\cref{eq:al}). The complexity of this step is $\mathcal{O}(mb^2d)$, where $m$ corresponds to the number of frames $T$, and $b$ to the spatial tokens per frame ($H \times W$). In typical video generation scenarios, the spatial dimension is significantly larger than the temporal one ($b \gg m$), causing the $b^2$ term to dominate the computation and rendering a naive implementation inefficient.
To address this, we propose an online-entropy based FlashAttention algorithm that computes the softmax attention output and the entropy in a single pass, significantly reducing data movement between HBM and on-chip SRAM and improving computational efficiency in the Monarch factor $\bm{R}$ updating.
The forward and backward passes are provided in ~\cref{alg:flash_attention_entropy} and ~\cref{alg:flash_attention_entropy_bwd} (see Appendix), respectively.

\section{Experiments}
\label{sec:experiments}

\begin{table*}[ht]
    \centering
    \small
    \caption{Quality and efficiency comparison of our and other methods, including Full Attention~\cite{dao2023flashattention}, VSA~\cite{zhang2025vsa}, and VMoBA~\cite{wu2025vmoba}.
We verify the generalizability of our method across different architectures, parameter scales, and varying resolutions under the same tuning setting.
We report Vbench~\cite{huang2024vbench} metrics to evaluate quality, including Aesthetic Quality (AQ), Background Consistency (BC), Dynamic Degree (DD), Imaging Quality (IQ), and Subject Consistency (SC).
We also report the efficiency metrics, including attention sparsity, TFLOPs, and inference time.
$^{\oslash}$ denotes Training-Free setting. $^{\star}$ denotes extrapolation on longer sequences using the Wan2.1-1.3B~\cite{wan2025wan} model tuned at $61 \times 448 \times 832$ resolution.
We highlight the results of VMonarch if it is the \textbf{best} or \underline{second best} among efficient attention methods.}
    \label{table:exp_effectiveness_full}
    \setlength\tabcolsep{4.0pt}
    \begin{tabular}{c|c|l|ccccc|ccc}
        \toprule
        \multirow{2}{*}{\textbf{Model}} & \multirow{2}{*}{\textbf{Resolution}} & \multirow{2}{*}{\textbf{Kernel}} 
        & \multicolumn{5}{c|}{\textbf{Quality}} & \multicolumn{3}{c}{\textbf{Efficiency}} \\
        \cmidrule(lr){4-8} \cmidrule(lr){9-11} 
        & & & \texttt{AQ} $\uparrow$ & \texttt{BC} $\uparrow$ & \texttt{DD} $\uparrow$ & \texttt{IQ} $\uparrow$ & \texttt{SC} $\uparrow$ & {Sparsity} $\uparrow$ & {TFLOPs} $\downarrow$ &  {Time\ (s)}$\downarrow$ \\
        \midrule
        \multirow{8}{*}{Wan2.1 1.3B} & \multirow{8}{*}{61x448x832} & FullAttn$^{\oslash}$ & 65.44\% & 95.08\% & 69.44\% & 64.86\% & 91.77\% & - & 159.7 & 63.4 \\
        & & VMoBA$^{\oslash}$ & 42.00\% & 90.64\% & 88.89\% & 53.78\% & 82.00\% & 90.0\% & 75.8 & 71.7 \\
        & & VSA$^{\oslash}$ & 42.86\% & 94.96\% & 63.89\% & 55.04\% & 89.17\% & 90.0\% & 69.5 & 49.9 \\
        & & VMonarch$^{\oslash}$ & \textbf{63.65\%} & {\underline {94.93\%}} & 54.17\% & \textbf{64.46\%} & \textbf{92.43\%} & 87.5\% & 75.4 & \textbf{47.7} \\
        \cmidrule(l){3-11}
        & & FullAttn & 66.07\% & 95.75\% & 68.06\% & 64.57\% & 94.15\% & - & 159.7 & 63.4 \\
        & & VMoBA & 65.58\% & 96.22\% & 62.50\% & 66.70\% & 93.00\% & 90.0\% & 75.8 & 71.7 \\
        & & VSA & 64.46\% & 95.09\% & 61.11\% & 63.87\% & 92.99\% & 90.0\% & 69.5 & 49.9 \\
        & & VMonarch & \textbf{65.58\%} & {\underline {95.84\%}} & \textbf{62.50\%} & {\underline {64.49\%}} & \textbf{93.23\%} & 87.5\% & 75.4 & \textbf{47.7} \\
        \midrule
        \multirow{6}{*}{Wan2.1 1.3B$^{\star}$} & \multirow{3}{*}{61x720x1280} & FullAttn & 65.49\% & 96.03\% & 76.39\% & 65.15\% & 93.64\% & - & 758.9 & 246.8 \\
        & & VSA & 65.07\% & 95.86\% & 63.89\% & 65.69\% & 92.07\% & 90.0\% & 208.6 & 125.8 \\
        & & VMonarch & {\underline {64.39\%}} & \textbf{96.13\%} & \textbf{69.44\%} & \textbf{65.84\%} & \textbf{93.82\%} & 87.5\% & 243.3 & {\underline {136.8}} \\
        \cmidrule(l){2-11}
        & \multirow{3}{*}{141x448x832} & FullAttn & 64.27\% & 92.73\% & 63.89\% & 61.28\% & 92.63\% & - & 640.5 & 214.2 \\
        & & VSA & 63.04\% & 92.21\% & 66.67\% & 59.80\% & 89.71\% & 90.0\% & 184.2 & 112.6 \\
        & & VMonarch & {\underline {60.45\%}} & \textbf{92.25\%} & \textbf{69.44\%} & \textbf{59.92\%} & {\underline {88.86\%}} & 94.4\% & 170.2 & \textbf{102.7} \\
        \midrule
        \multirow{4}{*}{Wan2.1 14B} & \multirow{4}{*}{93x704x1280} & FullAttn$^{\oslash}$ & 67.42\% & 96.02\% & 66.67\% & 66.73\% & 93.52\% & - & 7903.8 & 2222.2 \\
        & & FullAttn & 67.49\% & 97.04\% & 61.11\% & 68.53\% & 95.23\% & - & 7903.8 & 2222.2 \\
        & & VSA & 66.40\% & 96.93\% & 59.72\% & 65.91\% & 94.05\% & 90.0\% & 2642.7 & 970.9 \\
        & & VMonarch & {\underline {65.91\%}} & \textbf{97.32\%} & {\underline {58.33\%}} & \textbf{66.61\%} & \textbf{95.68\%} & 92.0\% & 2670.5 & \textbf{969.2} \\
        \midrule
        \multirow{4}{*}{Wan2.2 5B} & \multirow{4}{*}{93x704x1280} & FullAttn$^{\oslash}$ & 64.66\% & 96.10\% & 72.22\% & 65.24\% & 93.58\% & - & 352.4 & 123.7 \\
        & & FullAttn & 64.99\% & 96.67\% & 62.50\% & 65.44\% & 94.56\% & - & 352.4 & 123.7 \\
        & & VSA & 64.82\% & 96.30\% & 61.11\% & 65.26\% & 93.29\% & 90.0\% & 204.4 & 103.0 \\
        & & VMonarch & \textbf{65.04\%} & \textbf{96.68\%} & {\underline {59.72\%}} & \textbf{66.69\%} & \textbf{94.20\%} & 92.0\% & 205.5 & \textbf{92.0} \\
        \bottomrule
    \end{tabular}
\end{table*}
\subsection{Implementation Setup}
% Model and Datasets.

\textbf{Base Model and Datasets.}
We conduct experiments on three base models: Wan2.1-1.3B, Wan2.1-14B, and Wan2.2-5B~\cite{wan2025wan}. 
To adapt the attention mechanism, we utilize the Wan14B-Syn-600k dataset~\cite{zhang2025vsa} for the 1.3B model, and the Wan2.2-Syn-32k dataset~\cite{zhang2025vsa} for the larger 14B and 5B models.
We compare our VMonarch with current state-of-the-art sparse attention-based methods in Video DiTs, including VSA~\cite{zhang2025vsa} and VMoBA~\cite{wu2025vmoba}, and choose FlashAttention-2~\cite{dao2023flashattention} as our full attention backend.

% Metrics
\textbf{Metrics.}
Following prior works ~\cite{yang2024cogvideox,wu2025vmoba}, we use  VBench~\cite{huang2024vbench} prompts after optimization for the evaluation of all models.
We adopt five assessment aspects from VBench to evaluate the video generation quality. Specifically, we choose Aesthetic Quality (AQ), Background Consistency (BC), Dynamic Degree (DD), Imaging Quality (IQ), and Subject Consistency (SC) as our evaluation metrics.
Furthermore, we use Peak Signal to Noise Ratio (PSNR) to evaluate the similarity between efficient attention and full attention generated videos.

\textbf{Definition of Sparsity.} VSA uses a top-k strategy with a fixed $90\%$ sparsity ratio, and VMoBA uses a top-p strategy with $0.25$ threshold; we approximate it to $90\%$ sparsity for reference. We estimate the sparsity of VMonarch as $1 -t\cdot\frac{T+HW}{THW} \approx 1-\frac{t}{T}$
~\footnote{Two Monarch factors $L\in\mathbb{R}^{m\times b\times b}$ and $R\in\mathbb{R}^{b\times m\times m}$, letting $n=m \times b$, result in $1-\frac{m+b}{n}$ sparsity to $n\times n$ matrices.}
, where $t$ is the Monarch iteration number and $THW$ is the latent size (with $HW \gg T$).

\textbf{Implementation details.}
% Hyperparameters / Implementation details.
We fine-tune all models using the AdamW optimizer with a learning rate of $1\!\times\!10^{-6}$ and a batch size of 8.
Specifically, for Wan2.1-1.3B, we fine-tune the DiT backbone for 1500 steps at a resolution of $61 \times 448 \times 832$. 
For Wan2.1-14B and Wan2.2-5B, we fine-tune them for 800 steps at a resolution of $93 \times 704 \times 1280$, utilizing sequence parallelism of size 8.
For VMonarch, we set the block size to the number of tokens in one frame, \emph{i.e.} $HW$, and clamp the $\bm{c}_{\bm{R}}$ value to $0.1$ for numerical stability and use two Monarch iterations by default.
For VSA and VMoBA, we follow their official default configurations.
We did not employ common training-free tricks such as hybridizing with Full Attention in layers, heads, or early denoising steps, which could potentially improve performance.
During inference, we apply the same attention kernel across 50 denoising timesteps, without using full attention for the initial steps, as in prior works~\cite{wu2025vmoba, xi2025sparse}. Except for the attention kernel, all other inference settings are the same. This ensures a fair comparison of end-to-end performance across different attention mechanisms.
For sequence length extension experiments, we evaluate the above fine-tuned models in spatial dimensions extension ($61 \times 720 \times 1280$) and temporal dimensions extension ($141 \times 448 \times 832$). VMoBA is omitted due to chunk-divisibility requirements. See ~\cref{sup:exp_detail} for details.

\begin{table}[t]
\centering
\caption{
    Ablation studies on training-free setting. In 'VM-T$n$-F$k$', T represents the number of iterations, and F denotes the size of the Monarch factor $b$ corresponding to the number of frames. 
    The $\dagger$ indicates the model variant without first frame computation. 
}
\label{tab:ablations_training_free}
\footnotesize % Use smaller font to fit in one column
\setlength{\tabcolsep}{2.5pt} % Reduce column spacing
\begin{tabular}{lccccccc}
\toprule
\multirow{2}{*}{Model} & \multicolumn{2}{c}{Similarity} & \multicolumn{5}{c}{Vbench Score (\%)} \\
\cmidrule(lr){2-3} \cmidrule(lr){4-8}
& PSNR$\uparrow$ & SSIM$\uparrow$ & AQ & BC & DD & IQ & SC \\
\midrule
Softmax & - & - & 65.44 & 95.08 & 69.44 & 64.86 & 91.77 \\
\midrule
VM-T1-F1            & 11.18 & 0.35 & 54.36 & 89.75 & 16.67 & 54.93 & 93.17 \\
VM-T2-F1$^\dagger$  & 11.65 & 0.40 & 63.39 & 92.25 & 29.17 & 67.76 & 88.46 \\
VM-T2-F1            & 12.59 & 0.43 & 63.65 & 94.93 & 54.17 & 64.46 & 92.43 \\
VM-T3-F1            & 12.21 & 0.41 & 63.89 & 95.01 & 50.00 & 65.12 & 93.55 \\
VM-T1-F2            & 11.21 & 0.35 & 55.27 & 89.80 & 20.83 & 56.44 & 93.09 \\
\bottomrule
\end{tabular}%
\end{table}

\subsection{Method Effectiveness}
% Quantitative Results. 
\textbf{Quantitative Results.}
We present the quantitative results of our Video Monarch Attention compared to other baseline methods in~\cref{table:exp_effectiveness_full}.
In the training-free setting, both our Monarch Attention and other sparse attention methods exhibit a performance drop compared to the original full attention. This is expected, as the base model has not been adapted to the inductive biases of efficient attention kernels.
However, we observe that Monarch Attention significantly outperforms other sparse attention methods under high sparsity ratios ($90\%$), which we attribute to its preservation of the global spatio-temporal structure inherent in video data.
We also note a significant decrease in the Dynamic Degree for Monarch Attention in the training-free setting. We hypothesize that this is due to the approximation error from the alternating minimization optimization of the Monarch matrix.
Nevertheless, VMonarch's performance on Dynamic Degree is fully restored after minimal fine-tuning for 1500 steps, even surpassing the original full attention, while also outperforming other sparse attention-based models in Subject Consistency and Aesthetic Quality.
We evaluate the zero-shot generalization of the fine-tuned model with spatial and temporal sequence length extension. 
VMonarch demonstrates remarkable generalization capabilities. 
For temporal extension to $141 \times 448 \times 832$, it achieves results comparable to the original model while being superior to VSA. 
In the spatial extension setting of $61 \times 720 \times 1280$, our method again performs exceptionally well, outperforming both VSA and full attention on several key metrics, including Background Consistency, Dynamic Degree, Imaging Quality, and Subject Consistency, which demonstrates the powerful expressive capability of VMonarch.
VMonarch also maintains comparable or even superior performance to other methods across larger parameter scales and extended sequence lengths, verifying its strong generalizability.
% Qualitative Results. (Some Visualizations)

\begin{figure}[t]
  \centering
  \includegraphics[page=1,width=\linewidth]{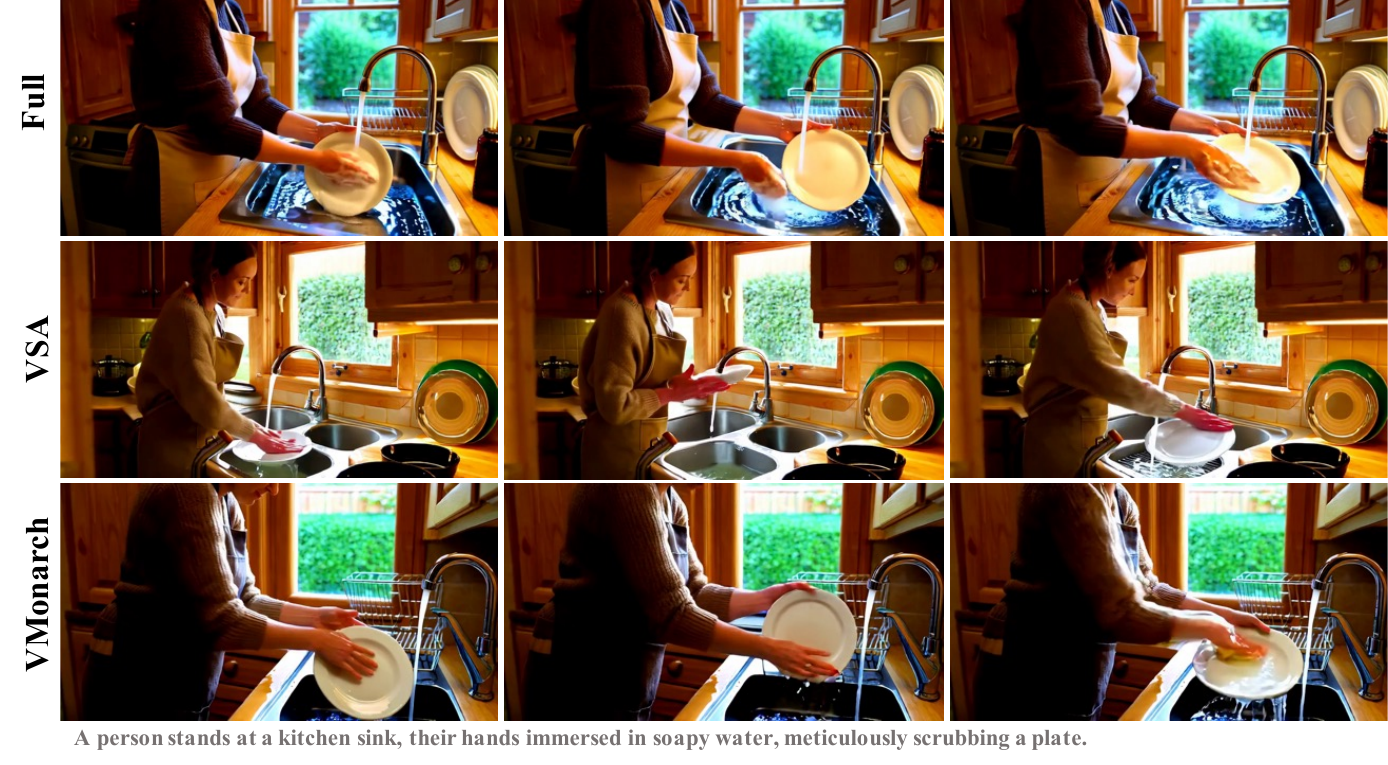}
  \includegraphics[page=1,width=\linewidth]{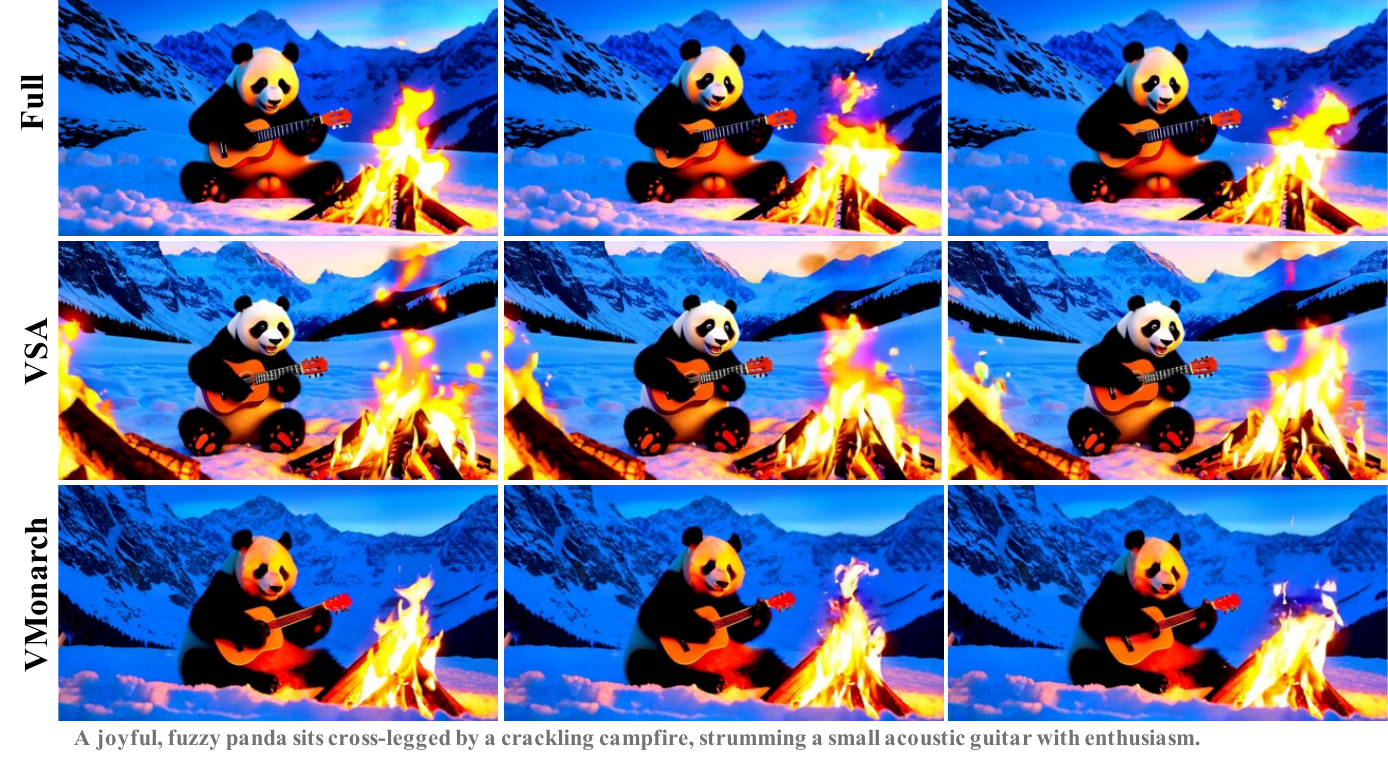}
  \includegraphics[page=1,width=\linewidth]{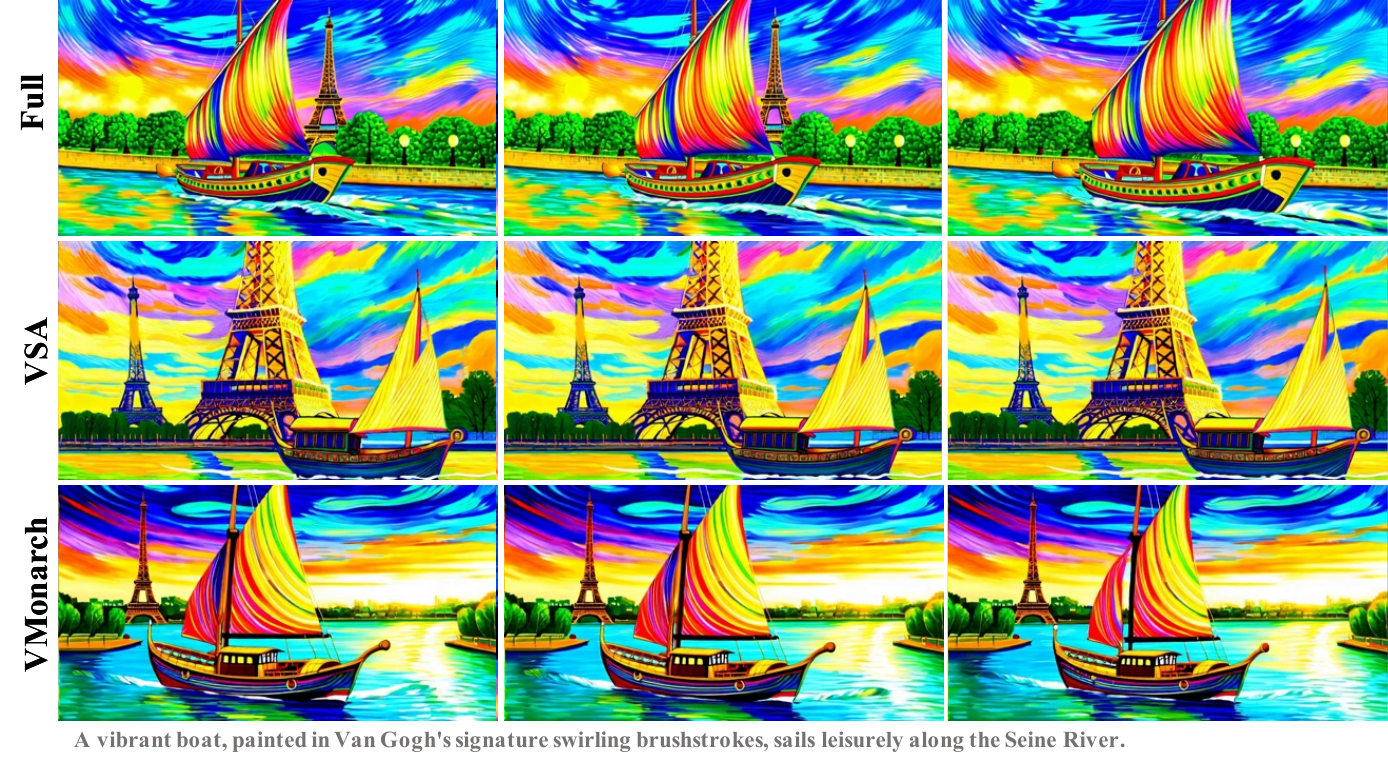}
  \caption{Qualitative samples of our models. We compare the generation quality of the tuned models between full attention~\cite{dao2023flashattention}, Video Sparse Attention (VSA)~\cite{zhang2025vsa}, and our VMonarch Attention.}
  \label{fig:qualitative_results}
  \vspace{-1em}
\end{figure}

\textbf{Qualitative Results.}
As shown in~\cref{fig:qualitative_results}, we present qualitative samples generated by Wan2.1-1.3B fine-tuned under identical settings. The results demonstrate that VMonarch generates videos with a quality comparable to that of full attention. In contrast, VSA occasionally produces incoherent content. This suggests that VMonarch better captures the spatio-temporal priors inherent in video generation, leading to more robust and consistent results.

\subsection{Method Efficiency} % Merge into Effectiveness part?
\begin{figure}[t]
    \centering
    \includegraphics[width=\linewidth]{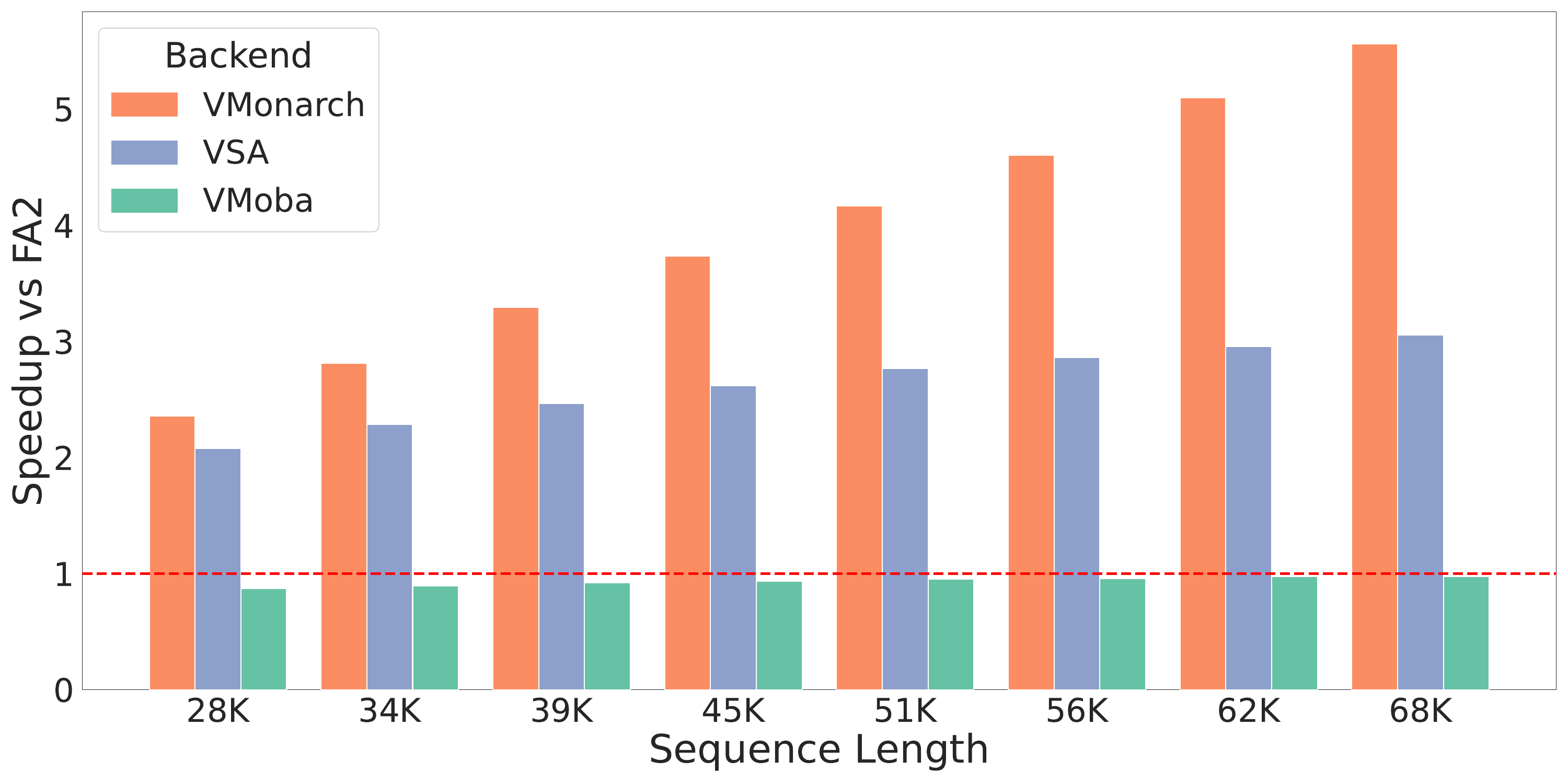}
    \caption{The kernel speedup ratio over Flash Attention 2 (FA2)~\cite{dao2023flashattention} under varying sequence lengths. 
    Our VMonarch demonstrates superior efficiency, achieving over 2$\times$ speedup at 28k tokens and over 5$\times$ speedup at 62k tokens compared to FA2, surpassing other sparse attention methods at $90\%$ sparsity.
    }
    \label{fig:speed_benchmark}
    \vspace{-2mm}
\end{figure}

As shown in~\cref{table:exp_effectiveness_full}, we evaluate the efficiency of different attention mechanisms in terms of FLOPs and end-to-end inference time.
The FLOPs of VMoBA is estimated at an equivalent 90\% sparsity for comparison.
Our VMonarch utilizes an estimated sparsity $(1 - t\cdot\frac{T+HW}{THW})$, where $t$ is the iteration number (set to 2), $(T,H,W)$ is the latent token shape.
In the $61 \times 448 \times 832$ setting, VMonarch reduces TFLOPs by 53\% and inference time by 25\% compared to full attention, even surpassing VSA which has lower TFLOPs, exhibiting superior practical efficiency from VMonarch's structured computation.
This efficiency advantage becomes more pronounced in long-sequence scenarios. For temporal extension to $141\times448\times832$, VMonarch achieves a 2.1$\times$ speedup over full attention and is also 9\% faster than VSA.
As shown in ~\cref{fig:speed_benchmark}, we also benchmark the kernel-level speed, gradually increasing the token number from 30k to 60k by varying the number of latent temporal dimensions from 20 to 48 under $(28,52)$ spatial size. 
The Monarch matrix size is set with spatial-temporal structure; VSA and VMoBA's block number is set to match the sparsity ratio of $90\%$.
Our VMonarch demonstrates significant efficiency gains, achieving over 2$\times$ speedup compared to FA2 at 34k tokens and over 5$\times$ speedup at 62k tokens.
% We also observe that block layouts critically affect the efficiency of sparse attention methods; ~\emph{i.e.}, small 3D cube layouts or Top-P strategy under a uniform distribution can lead to significant efficiency drops for VMoBA, as detailed in the~\cref{sup:detailed_speed_benchmark}. 
Furthermore, VMonarch achieves an 8$\times$ speedup over the naive implementation when optimized with Online-Entropy Flash Attention, with details in ~\cref{sup:kernel_optimization}.

\subsection{Ablation Studies}

% Different components Monarch Attention %
\textbf{Number of iterations.}
We investigate the impact of the number of iterations on Monarch Attention's performance. 
In the training-free setting, as shown in~\cref{tab:ablations_training_free}, increasing the iterations from 1 to 3 gradually improves most VBench metrics, with the Dynamic Degree peaking at two iterations.
As shown in~\cref{tab:ablations_tuned}, after fine-tuning, more iterations lead to a lower validation loss, indicating a stronger fitting capability but at the cost of increased computation. Extending the number of iterations up to 7 yields no significant gains while reducing efficiency.
Notably, the 2-iteration setting strikes an optimal balance between effectiveness and efficiency, outperforming the 3-iteration setting on Dynamic Degree with less computational overhead. Therefore, we adopt two Monarch iterations as our default.

\textbf{Monarch matrix size.}
% We study the impact of the Monarch matrix size $b$ on performance. 
We set the Monarch matrix size $b$ to the number of tokens in a single frame by default, ~\emph{i.e.}, $b=HW$, to align with the natural structure of video data. We compare this with a larger block size spanning two frames ($b=2HW$).
As shown in~\cref{tab:ablations_training_free} and~\cref{tab:ablations_tuned}, under the training-free setting, while the larger block size ($b=2F$) shows a slight improvement on metrics, it introduces visual temporal inconsistency artifacts, with abrupt changes occurring every two frames. This is due to a structural mismatch between the matrix size and the video's spatial-temporal structure. 
This partitioning-induced artifact is not resolved by fine-tuning. The model with $b=2F$ results in a higher validation loss, reduced Dynamic Degree and Imaging Quality. This underscores the importance of aligning the Monarch block structure with the intrinsic spatial-temporal structure of video data.

%% Recomputation of the first frame
\begin{table}[t]
\centering
\caption{Ablation studies on fine-tuned Video Monarch Attention models under different configurations. 
In 'VM-T$n$-F$k$', T represents iterations numbers, and F denotes the frame number corresponding to the Monarch Matrix $b$.
$^\dagger$ omits the recompute strategy for the first frame.
$\ddagger$ means leveraging attention sink by averaging.
}
\label{tab:ablations_tuned}
\small % Use a smaller font to fit in one column
\setlength{\tabcolsep}{3pt} % Reduce column spacing
\begin{tabular}{lcccccc}
\toprule
\multirow{2}{*}{Model} & \multicolumn{5}{c}{Vbench Score (\%)} & \multicolumn{1}{c}{Val} \\
\cmidrule(lr){2-6}
& AQ & BC & DD & IQ & SC & Loss \\
\midrule
Softmax & 65.72 & 94.85 & 75.00 & 64.46 & 92.59 & 1.09 \\
\midrule
VM-T1-F1                    & 66.53 & 95.82 & 59.72 & 65.28 & 92.69 & 1.27 \\
VM-T2-F1                    & 64.84 & 95.28 & 68.06 & 63.83 & 93.29 & 1.25 \\
VM-T3-F1                    & 66.22 & 96.12 & 56.94 & 65.72 & 94.06 & 1.24 \\
VM-T5-F1                    & 65.63 & 96.16 & 45.83 & 67.91 & 94.94 & 1.57 \\
VM-T7-F1                    & 64.39 & 96.71 & 36.11 & 65.51 & 94.61 & 1.55 \\
VM-T1-F2                    & 66.90 & 96.61 & 55.56 & 65.05 & 94.44 & 1.37 \\
VM-T2-F1$^\dagger$          & 63.22 & 92.19 & 69.44 & 63.63 & 87.36 & 1.27 \\
VM-T2-F1$^\ddagger$   & 63.85 & 94.94 & 45.83 & 67.40 & 93.83 & 1.78 \\
\bottomrule
\end{tabular}%

\vspace{-1em}
\end{table}
\textbf{First-frame recomputation.}
We analyze the effectiveness of our proposed first-frame recomputation strategy in mitigating oversmoothing in the first frame.
Under training-free settings, as shown in~\cref{tab:ablations_training_free}, omitting the first-frame computation results in a significant drop in PSNR and SSIM, as well as almost all VBench metrics, indicating a substantial deviation from the full attention output. 

To further isolate the impact on the initial frame, we additionally calculate the PSNR and SSIM on the first frame only. With our recomputation strategy, the first frame achieves a PSNR of 12.43 and an SSIM of 0.42. In contrast, without recomputation, these scores drop to 10.42 and 0.34, respectively, which underscores the necessity of the recomputation strategy for preserving first-frame quality.

As shown in~\cref{tab:ablations_tuned}, we evaluate our first-frame recomputation strategy by comparing it against two alternatives under the fine-tuning setting: one that omits the recomputation ($^\dagger$), and another that uses an attention sink mechanism ($\ddagger$).
The latter approach ($\ddagger$) computes the output by averaging the standard Monarch Attention with full attention to the first frame, defined as $O = (\text{Softmax}(Q, K_{F1}, V_{F1}) + \text{MonarchAttention}(Q, K, V))/2$, where $K_{F1}$ and $V_{F1}$ are the key and value from the first frame.
Compared to both alternatives, our recomputation strategy demonstrates superior performance. The attention sink approach ($\ddagger$) exhibited convergence issues during training, leading to a higher validation loss and degraded performance in Dynamic Degree and Subject Consistency. Our default strategy achieves a lower validation loss and higher scores across both image quality (Aesthetic Quality, Imaging Quality) and stability metrics (Background Consistency, Subject Consistency) compared to the variant without recomputation ($^\dagger$).

\section{Conclusion}
\label{sec:conclusion}
 We overcome the quadratic complexity bottleneck in DiT-based video generation by introducing VMonarch, a sub-quadratic attention mechanism that leverages an expressive class of Monarch matrices with spatial-temporal structure. Furthermore, we introduce a first-frame recomputation strategy to mitigate initial frame artifacts caused by the over-smoothing issue during the iterative minimization of Monarch matrices. We also develop a custom GPU kernel that merges FlashAttention with online entropy computation, enabling VMonarch to process video sequences efficiently with fast updates of Monarch matrices. Extensive experiments demonstrate that VMonarch achieves a superior trade-off between quality and efficiency, maintaining generation quality comparable or superior to full attention on VBench while delivering more than $5 \times$ attention kernel speedup and over a $17\times$ reduction in attention FLOPs for long video sequences.
{
    \small
    \bibliographystyle{ieeenat_fullname}
    \bibliography{main}
}

% WARNING: do not forget to delete the supplementary pages from your submission 
\clearpage
\setcounter{page}{1}
\maketitlesupplementary
\appendix

\section{Python code of MonarchAttention}\label{sec:code}
In this section, we show the Python-like code of MonarchAttention~\cite{yaras2025monarchattention} in ~\cref{fig:ma_code}.
\begin{figure}[h]
\begin{minted}[frame=lines, framesep=1.5mm, fontsize=\footnotesize, baselinestretch=1.1]{python}
def R_update(
    aR, # (m, b, d)
    cR, # (m, b)
    Kb, # (m, b, d)
    ): # Computes R, aL, cL from aR, cR
    R = softmax(bmm(aR, Kb.transpose(1, 2)) \
        / cR[:, :, None], dim=2)
    cL = sum(R * log(R), dim=2).transpose(0, 1)
    aL = bmm(R, Kb).transpose(0, 1)
    return aL, cL, R
\end{minted}
\vspace{-2.5em}
\begin{minted}[frame=lines, framesep=1.5mm, fontsize=\footnotesize, baselinestretch=1.1]{python}
def L_update(
    aL, # (b, m, d)
    cL, # (b, m)
    Qb, # (b, m, d)
    ): # Computes L, aR, cR from aL, cL
    L = softmax(bmm(Qb, aL.transpose(1, 2)) \
        - cL[:, None, :], dim=2)
    cR = sum(L, dim=1).transpose(0, 1)
    aR = bmm(L.transpose(1, 2), Qb).transpose(0, 1)
    return aR, cR, L
\end{minted}
\vspace{-2.5em}
\begin{minted}[frame=lines, framesep=1.5mm, fontsize=\footnotesize, baselinestretch=1.1]{python}
def monarch_attention(Q, K, V, m, b, T): 
    # Q, K, V: (N, d), m * b = N
    # T > 0: number of steps
    Qb = Q.reshape(m, b, d).transpose(0, 1)
    Kb = K.reshape(m, b, d)
    Vb = V.reshape(m, b, d)
    aR = Q.reshape(m, b, d)
    cR = ones(m, b)

    for t in range(T):
        aL, cL, R = R_update(aR, cR, Kb)
        aR, cR, L = L_update(aL, cL, Qb)

    y = bmm(R, Vb).transpose(0, 1)
    O = bmm(L, y).transpose(0, 1)
    O = O.reshape(N, d)
    return O
\end{minted}
\caption{Python-like code of MonarchAttention.}
\label{fig:ma_code}
\end{figure}

\section{Optimized Entropy Flash Attention}
The \textbf{online softmax algorithm} computes the softmax function over a vector $\mathbf{x} = [x_1, \dots, x_N]$ in a single pass without storing the entire vector. This is achieved by maintaining running statistics. For numerical stability, the algorithm uses the shift-invariant property of softmax with a running maximum $m = \max_{k}\{x_k\}$.
After processing $i$ elements, it maintains a state tuple $(m_i, S_i)$, where $m_i = \max_{k=1}^{i} \{x_k\}$ is the running maximum and $S_i = \sum_{k=1}^{i} e^{x_k-m_i}$ is the denominator sum normalized by $m_i$. When a new element $x_{i+1}$ arrives, the state is updated. The new maximum is $m_{i+1} = \max(m_i, x_{i+1})$. If the maximum changes ($m_{i+1} > m_i$), the previous sum $S_i$ must be rescaled by a factor $\alpha = e^{m_i-m_{i+1}}$. The general update for $S_i$ is:
\begin{equation}
S_{i+1} = S_i \cdot e^{m_i-m_{i+1}} + e^{x_{i+1}-m_{i+1}}
\end{equation}

We extend the online softmax algorithm to \textbf{online entropy}, which computes the Shannon entropy $H(p) = -\sum p_j \log p_j$ in a single pass. The entropy can be expressed in terms of the final statistics of the online softmax:

\begin{align}
H(p) &= -\sum_{j=1}^{N} p_j \log p_j \\
&= -\sum_{j=1}^{N} p_j \left( (x_j - m_N) - \log S_N \right) \\
&= \log S_N - \frac{1}{S_N} \sum_{j=1}^{N} e^{x_j-m_N} (x_j - m_N)
\end{align}
To compute this online, we introduce a third running statistic, $L_i$, which is the sum of logits weighted by their unnormalized probabilities: $L_i = \sum_{k=1}^{i} e^{x_k-m_i} (x_k - m_i)$.

We now present the complete update for the state $(m_i, S_i, L_i)$ upon receiving a new element $x_{i+1}$.
First, find the new maximum $m_{i+1} = \max(m_i, x_{i+1})$.

If the maximum does not change ($m_{i+1} = m_i$), the updates are additive:
\begin{align}
S_{i+1} &= S_i + e^{x_{i+1}-m_{i+1}} \\
L_{i+1} &= L_i + e^{x_{i+1}-m_{i+1}} (x_{i+1} - m_{i+1})
\end{align}

If the maximum increases ($m_{i+1} > m_i$), the previous sums must be rescaled before adding the new term. Let $\alpha = e^{m_i-m_{i+1}}$. The updates are:
\begin{align}
S_{i+1} &= S_i \cdot \alpha + e^{x_{i+1}-m_{i+1}} \\
L_{i+1} &= L_i \cdot \alpha + S_i \cdot \alpha \cdot (m_i - m_{i+1}) \nonumber \\ 
         &\quad + e^{x_{i+1}-m_{i+1}} (x_{i+1} - m_{i+1})
\end{align}

Noting that $\log(\alpha) = m_i - m_{i+1}$ and letting $p_{i+1} = e^{x_{i+1}-m_{i+1}}$, the update for $L_{i+1}$ can be expressed more compactly in a form analogous to the entropy formula itself:
\begin{equation}
L_{i+1} = L_i \cdot \alpha + S_i \cdot \alpha \log(\alpha) + p_{i+1} \log(p_{i+1})
\end{equation}

After processing all $N$ elements, the final state is $(m_N, S_N, L_N)$. The Shannon entropy is then calculated as:
\begin{equation}
H(p) = \log S_N - \frac{L_N}{S_N}
\end{equation}
% This method computes the exact entropy in a single, numerically stable pass.

\begin{algorithm}[!t]
\small 
\caption{Flash-{\color{blue}{Entropy}}-Attention Backward}
\label{alg:flash_attention_entropy_bwd}
\begin{algorithmic}[1]
\Require $\mathbf{Q}, \mathbf{K}, \mathbf{V}, \mathbf{O}, \mathbf{dO} \in \mathbb{R}^{N \times d}$, $L$, {\color{blue}{${H}, {dH} \in \mathbb{R}^{N}$}}.
\Ensure $\mathbf{dQ}, \mathbf{dK}, \mathbf{dV} \in \mathbb{R}^{N \times d}$.
\State $T_r \gets \lceil N/B_r \rceil$, $T_c \gets \lceil N/B_c \rceil$.
\State Compute $D \in \mathbb{R}^{N}$ as $\mathrm{rowsum}(\mathbf{dO} \odot \mathbf{O})$.
\State Initialize $\mathbf{dQ} \gets \mathbf{0}_{N \times d}$ in HBM.
\For{$j = 1 \to T_c$}
  \State Load $\mathbf{K}_j, \mathbf{V}_j$ to SRAM.
  \State Initialize $\mathbf{dK}_j, \mathbf{dV}_j \gets \mathbf{0}_{B_c \times d}$.
  \For{$i = 1 \to T_r$}
    \State Load $\mathbf{Q}_i, \mathbf{O}_i, \mathbf{dO}_i, L_i, D_i,$ {\color{blue}{${H}_i, d{H}_i$}} to SRAM.
    \State $\mathbf{S}_{ij} \gets \mathbf{Q}_i \mathbf{K}_j^T$.
    \State $\mathbf{P}_{ij} \gets \exp(\mathbf{S}_{ij} - L_i)$.
    \State $\mathbf{dV}_j \gets \mathbf{dV}_j + \mathbf{P}_{ij}^T \mathbf{dO}_i$.
    \State $\mathbf{dP}_{ij} \gets \mathbf{dO}_i \mathbf{V}_j^T$.
    \State $\mathbf{dS}_{ij} \gets \mathbf{P}_{ij} \odot (\mathbf{dP}_{ij} - D_i)$. \Comment{Standard Gradient}
    \State {\color{blue}{$\mathbf{dS}_{ij} \gets \mathbf{dS}_{ij} - {dH}_i \odot \mathbf{P}_{ij} \odot (\mathbf{S}_{ij} - L_i + {H}_i)$}} 
    \State Load $\mathbf{dQ}_i$ from HBM.
    \State $\mathbf{dQ}_i \gets \mathbf{dQ}_i + \mathbf{dS}_{ij} \mathbf{K}_j$.
    \State $\mathbf{dK}_j \gets \mathbf{dK}_j + \mathbf{dS}_{ij}^T \mathbf{Q}_i$.
    \State Write $\mathbf{dQ}_i$ to HBM.
  \EndFor
  \State Write $\mathbf{dK}_j, \mathbf{dV}_j$ to HBM.
\EndFor
\end{algorithmic}
\end{algorithm}

\section{Backward pass of Flash-Entropy-Attention}
We provide the backward pass algorithm of FlashAttention with Online Entropy in ~\cref{alg:flash_attention_entropy_bwd}.
However, the entropy term is not numerically stable; meanwhile, the gradient from entropy tends to dominate the gradient rather than the original attention output, which caused training instability.
Therefore, we do not backpropagate through the entropy calculation in practice. We still provide the full backward algorithm here for completeness.

\section{Experiments details}\label{sup:exp_detail}
We select Wan2.1-1.3B~\cite{wan2025wan} as our primary base model for training-free and post-training experiments, while also extending our verification to the larger Wan2.1-14B model~\cite{wan2025wan} and the architecturally different Wan2.2-5B model~\cite{wan2025wan}.
For the Wan2.1-1.3B post-training, we tune the model with a $61 \times 448 \times 832$ input size for 1500 steps on the Fastvideo Wan14B-Syn-600k dataset~\cite{zhang2025vsa}, using a learning rate of $1 \times 10^{-6}$ and a batch size of 8. 
For ablation studies on this model, we fine-tune for 500 steps with the same learning rate and batch size, calculating validation loss on 32 samples split from the training dataset.
Regarding the Wan2.1-14B and Wan2.2-5B models, we conduct fine-tuning on the Wan2.2-Syn-32k dataset~\cite{zhang2025vsa} with a resolution of $93 \times 704 \times 1280$. To handle the increased computational load, we employ sequence parallelism with a degree of 8. Regarding the sparse attention baselines, we follow their default setting. For VSA~\cite{zhang2025vsa}, we employ a sparsity annealing schedule that gradually increases the sparsity ratio to $90\%$, and use its Triton version as the CUDA version is incompatible with our hardware.
For VMoBA, we adopt its default cyclic partitioning strategy, alternating between 1D, 2D, and 3D attention blocks. We set the chunk sizes to 4, (7, 13), and (4, 7, 13), respectively, and use a top-$p$ ($0.25$) strategy for block selection for both training and inference.
We test all models on VBench-Long~\cite{huang2024vbench} prompts with the same setting: 50 inference steps, guidance scale 5.0, and flow shift 3.0, the negative prompt is "Bright tones, overexposed, static, blurred details, subtitles, style, works, paintings, images, static, overall gray, worst quality, low quality, JPEG compression residue, ugly, incomplete, extra fingers, poorly drawn hands, poorly drawn faces, deformed, disfigured, misshapen limbs, fused fingers, still picture, messy background, three legs, many people in the background, walking backwards".

\section{Various Monarch Computation Budgets}
\begin{figure}[t]
    \centering
    \renewcommand{\thefigure}{A}
    \includegraphics[width=1.0\linewidth]{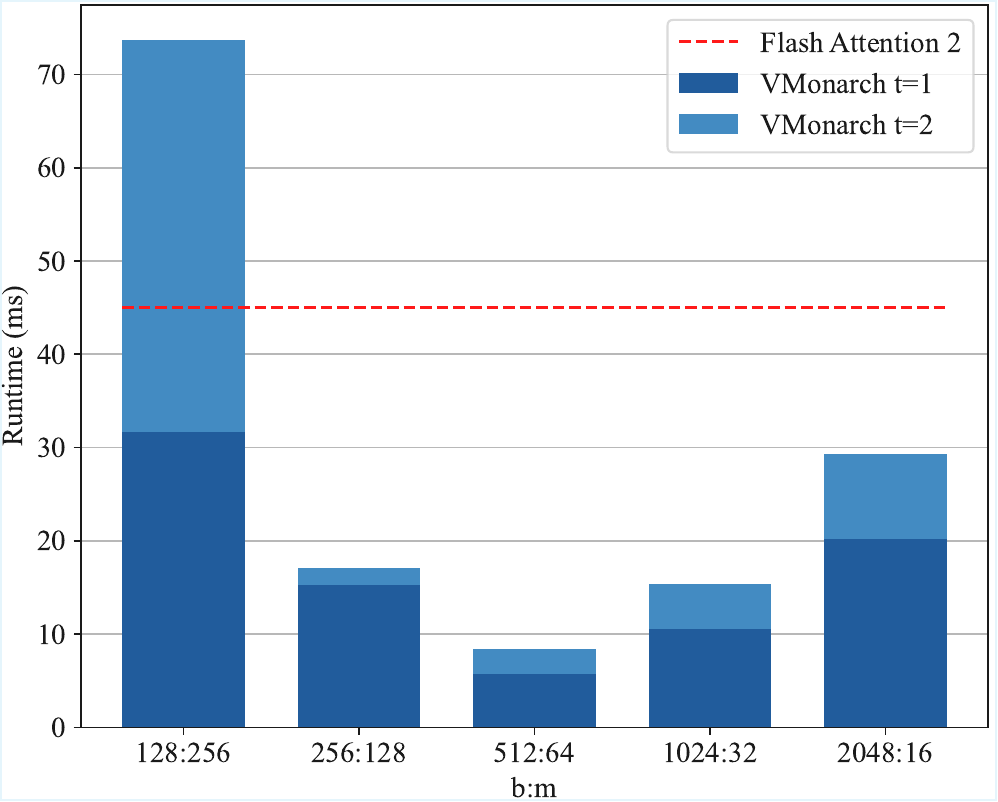}
    \vspace{-1.1em}
    \caption{Various computational budgets compared with FA2
    % under $2^{15}$ sequence length. 
    at a sequence length of $2^{15}$.
    In practice, $b$ is usually larger than $256$. }
    \label{fig:speed_benchmark_budget}
    \vspace{-1.5em}
\end{figure}
Our kernel supports flexible attention decomposition, and its computational budget can be adjusted by varying the monarch factor sizes $m,b$.
Theoretically, the computational complexity is minimized when $m=b=\sqrt{N}$, and the I/O complexity decreases as $b$ increases. As the I/O complexity decreases as $m$ decreases, the best $m$ lies in $[1, \sqrt N]$. 
As shown in~\cref{fig:speed_benchmark_budget}, we profile the kernel inference time with various computation budgets at a sequence length of 32k.
In the video modality, setting $b\!=\!1\text{F}$ (where $\text{F}\!=\!HW$) corresponds to a specific decomposition ($b:m	\approx 1024:32$), which can be further accelerated by reducing $b$ via spatial downsampling.

\section{Kernel Optimization of VMonarch}
\label{sup:kernel_optimization}
\begin{figure}[t]
    \centering
    \includegraphics[width=1.0\linewidth]{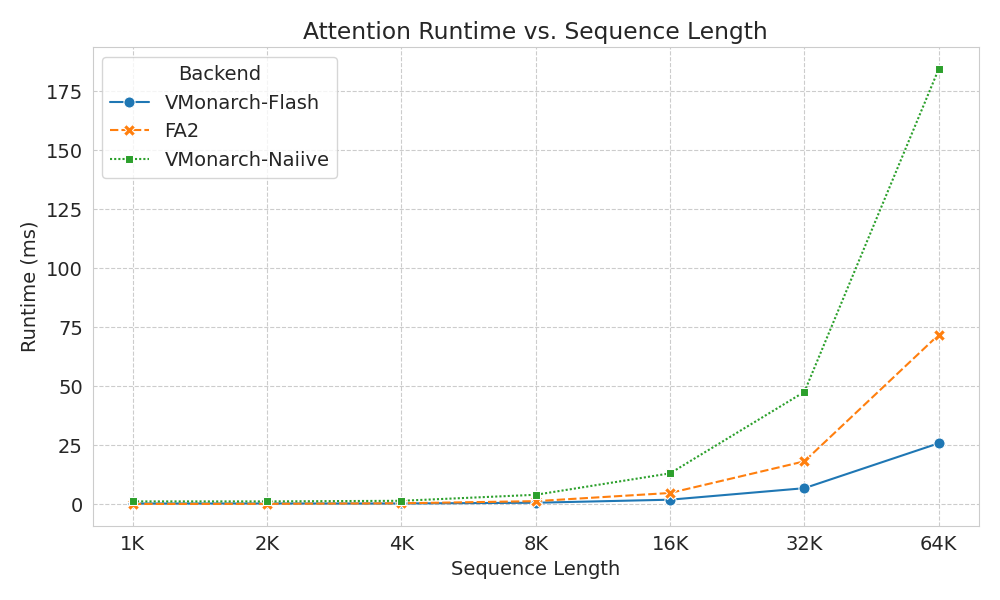}
    \caption{Speed comparison of FlashAttention2(FA2)~\cite{dao2023flashattention}, VMonarch-Naive, and our optimized VMonarch-Fast (Triton) implementation. After applying kernel optimization, VMonarch-Fast achieves an approximately $8 \times$  speedup over the naive implementation and surpasses the performance of FA2.}
    \label{fig:kernel_speedup}
\end{figure}

To further enhance the efficiency of our Video Monarch Attention (VMonarch), we developed a fused kernel. We benchmark the speed of our optimized kernel against a naive PyTorch implementation. 
The benchmark is conducted with a batch size of 1, 12 attention heads, and a head dimension of 64. The number of iterations for VMonarch is set to 2. We test on sequence lengths ranging from $2^{10}$ to $2^{16}$. The Monarch factor sizes $m$ and $b$ are set to $m=32$ and $b = \text{sequence length} / 32$.
As shown in~\cref{fig:kernel_speedup}, the optimized kernel provides a significant speedup, achieving approximately $8 \times$ speedup over the naive implementation and surpassing the performance of FlashAttention2 (FA2)~\cite{dao2023flashattention}.

\section{Detailed Speed and Memory Benchmarks}\label{sup:detailed_speed_benchmark}
\begin{figure}[ht]
    \centering
    \includegraphics[width=1.0\linewidth]{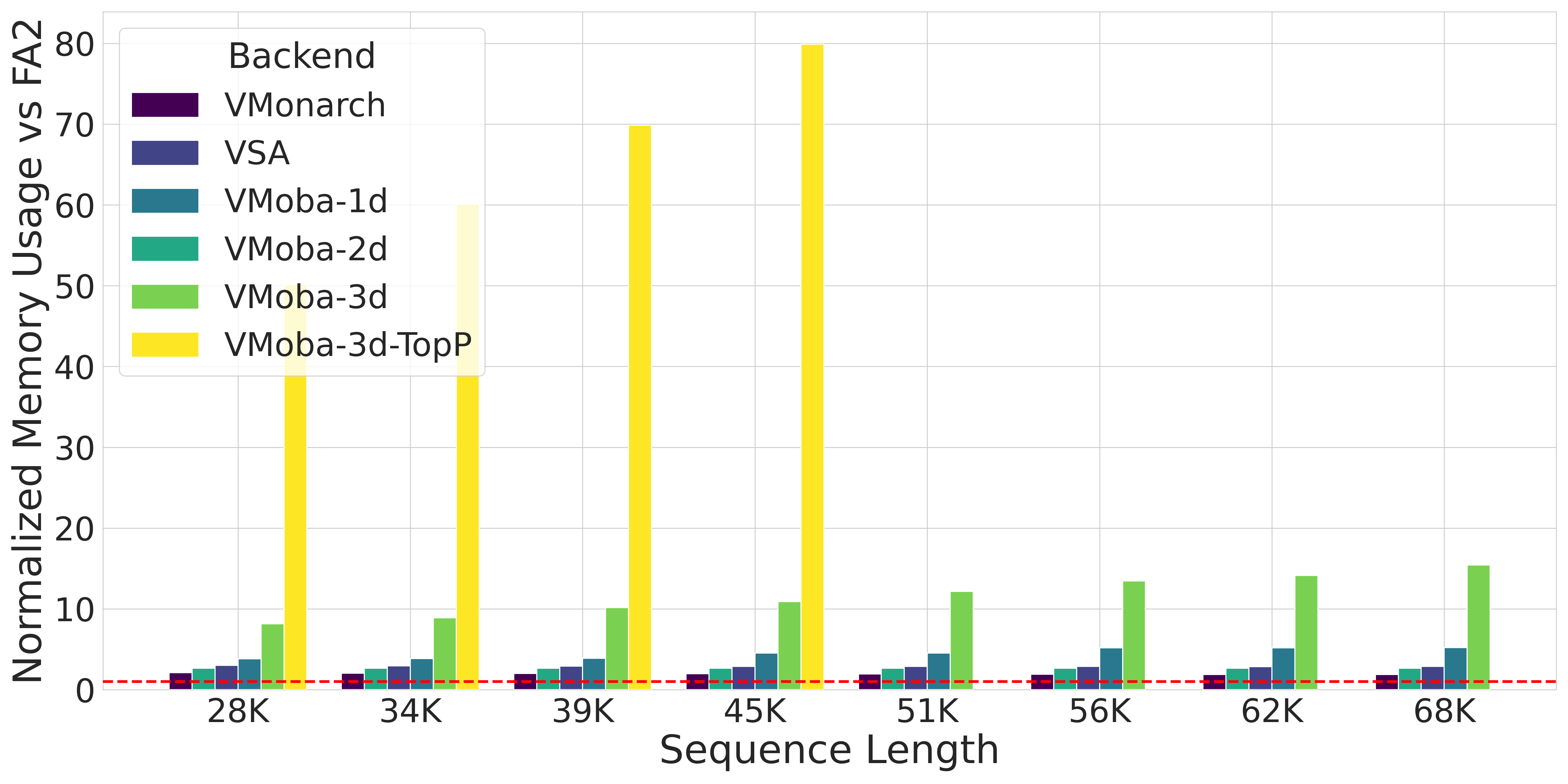}
    \caption{Memory benchmark of different attention mechanisms. We compare the relative memory usage against full attention.}
    \label{fig:detail_memory_benchmark}
\end{figure}

\begin{figure}[ht]
    \centering
    \includegraphics[width=1.0\linewidth]{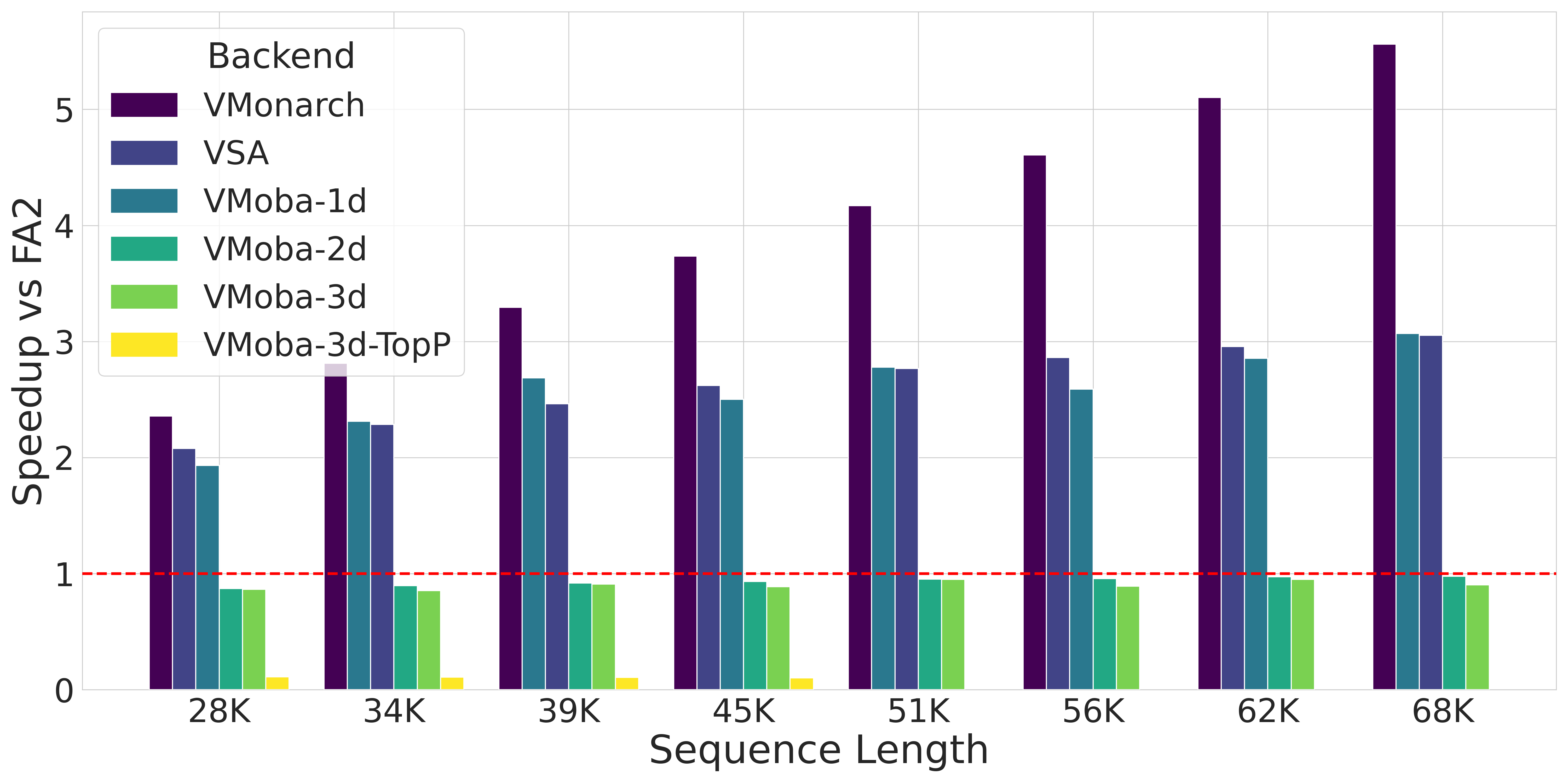}
    \caption{Speed benchmark of different attention mechanisms. We compare the relative speed against full attention.}
    \label{fig:detail_speed_benchmark}
\end{figure}

We benchmark the memory usage and speed of our Video Monarch Attention (VMonarch) against several baselines: Full Attention (FA2), Video Sparse Attention (VSA), and various VMoBA~\cite{wu2025vmoba} configurations (VMoBA-1d, VMoBA-2d, VMoBA-3d, and VMoBA-3d-TopP). The comparisons are conducted on different video lengths, with results presented in~\cref{fig:detail_memory_benchmark} and~\cref{fig:detail_speed_benchmark}. Our findings indicate that VMoBA encounters a significant memory overhead bottleneck at finer granularities. Furthermore, with random inputs (uniform distribution), the TopP strategy used in VMoBA-3d-TopP leads to a noticeable slowdown in speed and an increase in memory consumption.  We compare with FA2 as our kernel extends its Triton implementation. Our Online-Entropy Attention is also compatible with FA3, as both are based on the Online-Softmax algorithm.

\section{Detailed  Training-free comparison}
As shown in ~\cref{table:exp_effectiveness_psnr}, VMonarch outperforms VSA and VMoBA in PSNR and SSIM under the training-free setting, demonstrating that Monarch’s block-diagonal structure effectively aligns with the spatio-temporal priors of video.
\begin{table}[t]
    \centering
    % \vspace{-1em}
    \caption{Similarity comparison of our and other baseline methods. All methods are evaluated under the training-free setting with a resolution of $61\times448\times832$.}
    \small
    \label{table:exp_effectiveness_psnr}
    \setlength\tabcolsep{3.5pt}
    \begin{tabular}{c|cc}
    \toprule
    \multirow{2}{*}{\textbf{Method}} 
    & \multicolumn{2}{c}{\textbf{Similarity}} \\
    \cmidrule(lr){2-3} & PSNR $\uparrow$ & SSIM $\uparrow$ \\
    \midrule
    FlashAttention2 (FullAttn)~\cite{dao2023flashattention}    & - & - \\
    Video Sparse Attention (VSA)~\cite{zhang2025vsa}          & 9.71 & 0.32 \\
    Video MoBA (VMoBA)~\cite{wu2025vmoba}        & 10.61 & 0.30 \\
    Video Monarch (VMonarch)    & 12.59 & 0.43 \\
    \bottomrule
    \end{tabular}
    \vspace{-1.75em}
\end{table}

\section{Comparison of Attention Maps}
\begin{figure}[t]
    \centering
    \includegraphics[width=0.9\linewidth]{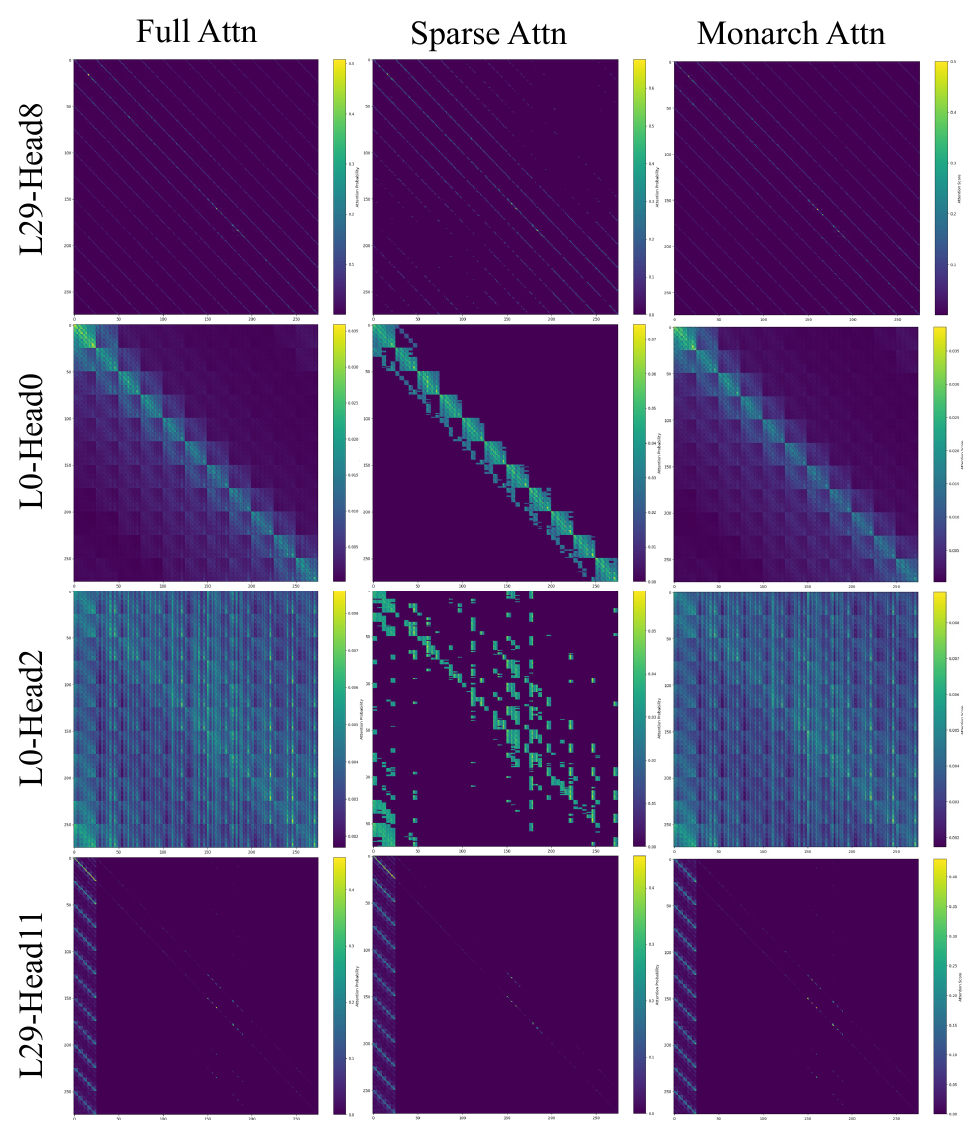}
    \caption{Attention map comparison among Full Attention, Sparse Attention, and Monarch Attention. }
    \label{fig:attention_map_comparison}
\end{figure}

We visualize the attention maps of Full Attention~\cite{dao2023flashattention}, Sparse Attention(corresponding to VSA~\cite{zhang2025vsa}), and Monarch Attention(corresponding to VMonarch) of different heads on Wanx1.3B~\cite{wan2025wan}. We downsample the attention maps to $300\times300$ for better visualization. 
As shown in~\cref{fig:attention_map_comparison}, Full Attention produces a dense attention map, while Sparse Attention generates a highly sparse attention map with many zero values.
Our VMonarch does not produce an overly sparse distribution and is much more continuous compared to VSA.

\section{Manual Control of Monarch Parameters}
\begin{table*}[]
\centering 
\caption{Ablation study on VMonarch Attention. We evaluate several variants by adjusting its iterative parameters, including the number of iterations and the manual control of intermediate terms. Metrics with a superscript '1' (e.g., $PSNR^{1}$) are calculated on the first frame only. The score is calculated as $\text{Score} = (\text{PSNR} + \text{PSNR}^{1})/200 + (\text{SSIM} + \text{SSIM}^{1})/2 + (1 -\text{LPIPS} + 1 - \text{LPIPS}^{1})/2 + \text{IQ} + \text{SC} + \text{OC}$, where IQ, SC, and OC denote Imaging Quality, Subject Consistency, and Overall Consistency, respectively. In the Method column, M$^{(t)}$ denotes the number of iterations $t$. +Re represents our recomputation mechanism. For the control of $\bm{c_R}$, +$\bm{c_R}^{*}$ indicates fixing it to 1 for all frames, +$\bm{c_R}^{1}$ for the first frame only, and +$\bm{c_R}^{0}$ means no fixing. For the control of $\bm{c_L}$, +$\bm{c_L}^{*}$ indicates fixing it to the initial expected value $\log(M)$ for all frames, where $M$ is the monarch factor size, +$\bm{c_L}^{1}$ for the first frame only, and +$\bm{c_L}^{0}$ means no fixing.
}
\label{tab:supl_control_crcl}
\resizebox{\textwidth}{!}{
\begin{tabular}{r|l|lllllllll|l}
\hline
Exp. & Method                      & PSNR $\uparrow$ & \multicolumn{1}{c}{SSIM $\uparrow$} & \multicolumn{1}{c}{LPIPS $\downarrow$} & $\text{PSNR}^{1}$ $\uparrow$ & $\text{SSIM}^{1}$ $\uparrow$ & $\text{LPIPS}^{1}$ $\downarrow$ & \multicolumn{1}{c}{IQ $\uparrow$} & \multicolumn{1}{c}{SC $\uparrow$} & \multicolumn{1}{c}{OC $\uparrow$} & \multicolumn{1}{c}{\textbf{Score} $\uparrow$} \\ \hline
1 & M$^{(1)}$                   & 11.01           & 0.21                                & 96.41\%                                & 10.21                              & 0.12                               & 90.44\%                               & 48.54\%                           & 87.00\%                           & 16.28\%                           & 2.19                                 \\ \hline
2 & M$^{(1)}$+$\bm{c_R}^{*}$+$\bm{c_L}^{1}$ & 10.66           & 0.23                                & 80.39\%                                & 11.21                              & 0.30                               & 51.54\%                               & 62.22\%                           & 87.53\%                           & 18.85\%                           & 3.11                                 \\ 
3 & M$^{(1)}$+$\bm{c_R}^{1}$+$\bm{c_L}^{1}$ & 10.66           & 0.23                                & 80.39\%                                & 11.21                              & 0.30                               & 51.54\%                               & 62.22\%                           & 87.53\%                           & 18.85\%                           & 3.11                                 \\
4 & M$^{(1)}$+$\bm{c_R}^{1}$+$\bm{c_L}^{*}$ & 11.01           & 0.18                                & 99.78\%                                & 10.29                              & 0.11                               & 95.63\%                               & 41.32\%                           & 85.87\%                           & 15.69\%                           & 1.98                                 \\
5 & M$^{(1)}$+$\bm{c_R}^{*}$+$\bm{c_L}^{*}$ & 11.01           & 0.18                                & 99.78\%                                & 10.29                              & 0.11                               & 95.63\%                               & 41.32\%                           & 85.87\%                           & 15.69\%                           & 1.98                                 \\
6 & M$^{(1)}$+$\bm{c_R}^{1}$+$\bm{c_L}^{0}$ & 11.01           & 0.21                                & 96.41\%                                & 10.21                              & 0.12                               & 90.44\%                               & 48.54\%                           & 87.00\%                           & 16.28\%                           & 2.19                                 \\
7 & M$^{(1)}$+$\bm{c_R}^{0}$+$\bm{c_L}^{1}$ & 10.66           & 0.23                                & 80.39\%                                & 11.21                              & 0.30                               & 51.54\%                               & 62.22\%                           & 87.53\%                           & 18.85\%                           & 3.11                                 \\ \hline
8 & M$^{(2)}$                   & 11.19           & 0.28                                & 52.44\%                                & 9.39                               & 0.21                               & 55.53\%                               & 72.17\%                           & 75.96\%                           & 19.47\%                           & 3.29                                 \\ \hline
9 & M$^{(2)}$+$\bm{c_R}^{*}$+$\bm{c_L}^{1}$ & 6.93            & 0.11                                & 78.38\%                                & 7.90                               & 0.16                               & 71.14\%                               & 48.53\%                           & {97.68\%}                  & 3.02\%                            & 2.42                                 \\
10 & M$^{(2)}$+$\bm{c_R}^{1}$+$\bm{c_L}^{1}$ & 6.94            & 0.11                                & 78.59\%                                & 7.78                               & 0.16                               & 71.70\%                               & 49.00\%                           & 97.88\%                           & 2.97\%                            & 2.41                                 \\
11 & M$^{(2)}$+$\bm{c_R}^{1}$+$\bm{c_L}^{*}$ & 10.71           & 0.24                                & 64.26\%                                & 8.68                               & 0.12                               & 58.73\%                               & 75.60\%                           & 90.48\%                           & 14.03\%                           & 3.12                                 \\
12 & M$^{(2)}$+$\bm{c_R}^{*}$+$\bm{c_L}^{*}$ & 8.12            & 0.17                                & 64.85\%                                & 6.96                               & 0.07                               & 59.14\%                               & 52.85\%                           & 96.08\%                           & 8.66\%                            & 2.73                                 \\
13 & M$^{(2)}$+$\bm{c_R}^{1}$+$\bm{c_L}^{0}$ & 11.13           & 0.27                                & 55.81\%                                & 9.14                               & 0.19                               & 55.01\%                               & 69.62\%                           & 76.55\%                           & {19.72\%}                  & 3.21                                 \\
14 & M$^{(2)}$+$\bm{c_R}^{0}$+$\bm{c_L}^{1}$ & 6.51            & 0.13                                & 80.88\%                                & 7.16                               & 0.18                               & 71.79\%                               & 49.18\%                           & 97.23\%                           & 1.51\%                            & 2.40                                 \\ \hline
15 & M$^{(2)}$+Re                & {11.55}  & {0.29}                       & {56.96\%}                       & 10.13                              & 0.27                               & 58.18\%                               & {73.06\%}                  & 91.26\%                           & 17.19\%                           & {3.44}                        \\ \hline
\end{tabular}
}
\end{table*}

To better understand the behavior of Monarch Attention, we conduct a detailed ablation study on its key components, with results presented in~\cref{tab:supl_control_crcl}. We analyze the impact of the number of iterations ($t$) and the manual control of the intermediate optimization terms, $\bm{c_L}$ and $\bm{c_R}$. For $\bm{c_L}$ and $\bm{c_R}$, we explore three strategies applied after each iteration: no control (using the updated value), fixing the value for the first frame only, and fixing the value for all frames. 

Our findings reveal several key insights. First, the number of iterations is crucial; increasing from one ($Exp.1$, Score: 2.19) to two ($Exp.8$, Score: 3.29) yields a substantial performance improvement. Second, while manually controlling $\bm{c_L}$ and $\bm{c_R}$ can boost the score in the single-iteration setting (e.g., $Exp.2$ achieves a score of 3.11), this benefit vanishes with two iterations. In the cases of two iterations ($Exp[8 \text{-}14]$), none of the manual control strategies outperform the baseline. This suggests that intervening in the optimization process may disrupt the stability of the alternating updates, leading to suboptimal performance.

In contrast, our proposed recomputation mechanism ($Exp.15$) provides a simple yet effective solution. It achieves the highest overall score of 3.44, outperforming all manual control variants and underscoring its effectiveness in enhancing generation quality without destabilizing the core optimization.

\section{Different Monarch Factorization methods}
\begin{figure}[h]
    \centering
    \includegraphics[width=1.0\linewidth]{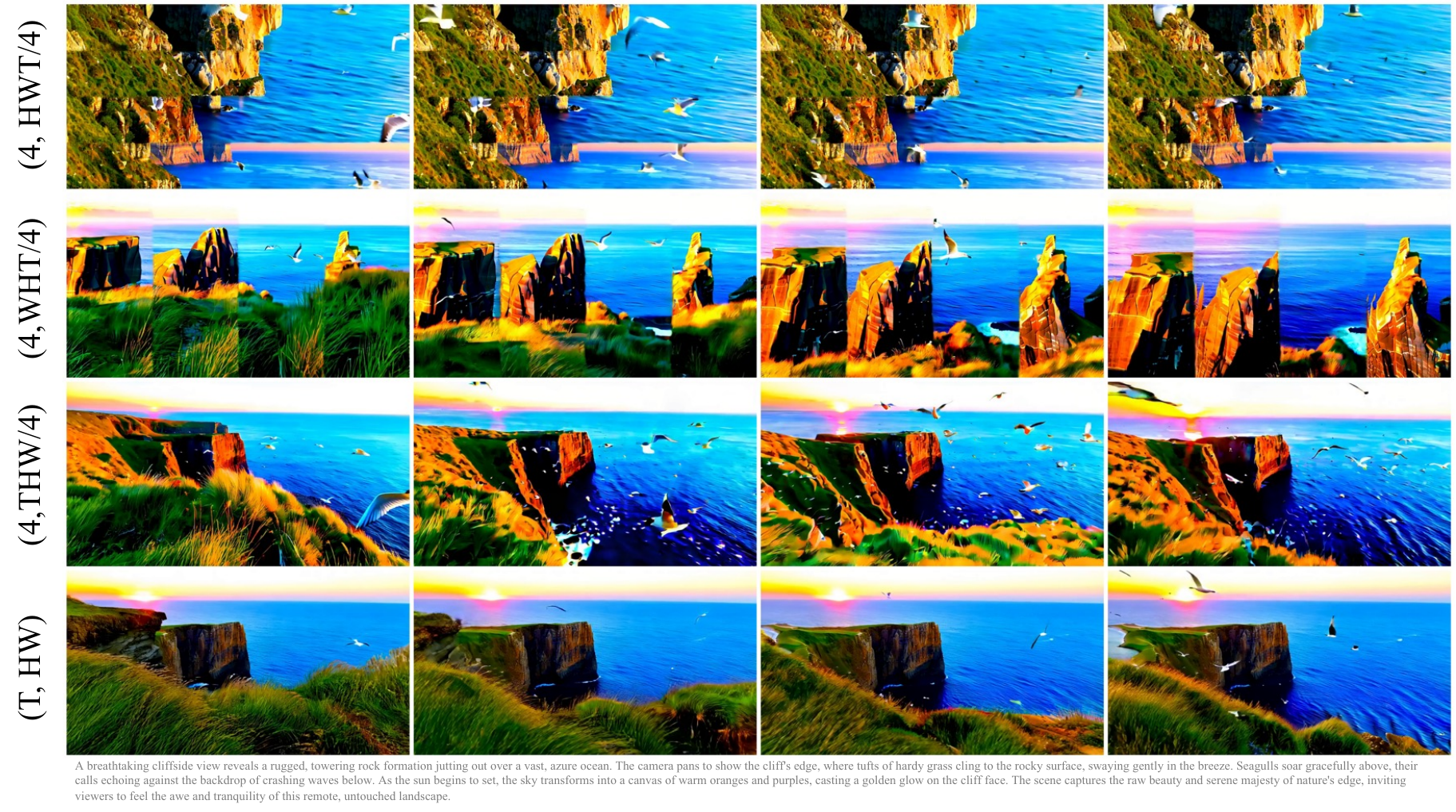}
    \caption{Comparison of different Monarch factorizations.}
    \label{fig:monarch_factorization_comparison}
    % \vspace{-3mm}
\end{figure}

Beyond the temporal-axis Monarch factorization method, we also explored alternative factorization strategies, such as spatial dimension factorization. Specifically, we grouped the video's spatial dimensions $H$ or $W$ with a grouping unit of 4, referred to as $(4, HWT/4)$ and $(4, WHT/4)$ respectively.
We conducted a simple comparison of different grouping methods in a zero-shot setting, with the results shown in~\cref{fig:monarch_factorization_comparison}.
Grouping along the $H$ or $W$ dimensions results in significant spatial discontinuities.
Additionally, we experimented with larger temporal blocks for grouping, such as dividing the sequence into four temporal segments, referred to as $(4, THW/4)$. We observed that this grouping method leads to substantial temporal jumps between blocks, although temporal consistency within blocks remains stable. Our default setting $(T, HW)$ provides the best balance between spatial and temporal coherence.

\section{More Visual Results}
\begin{figure*}[ht]
    \centering
    % Row 1
    \begin{subfigure}{0.45\linewidth}
        \includegraphics[width=\linewidth]{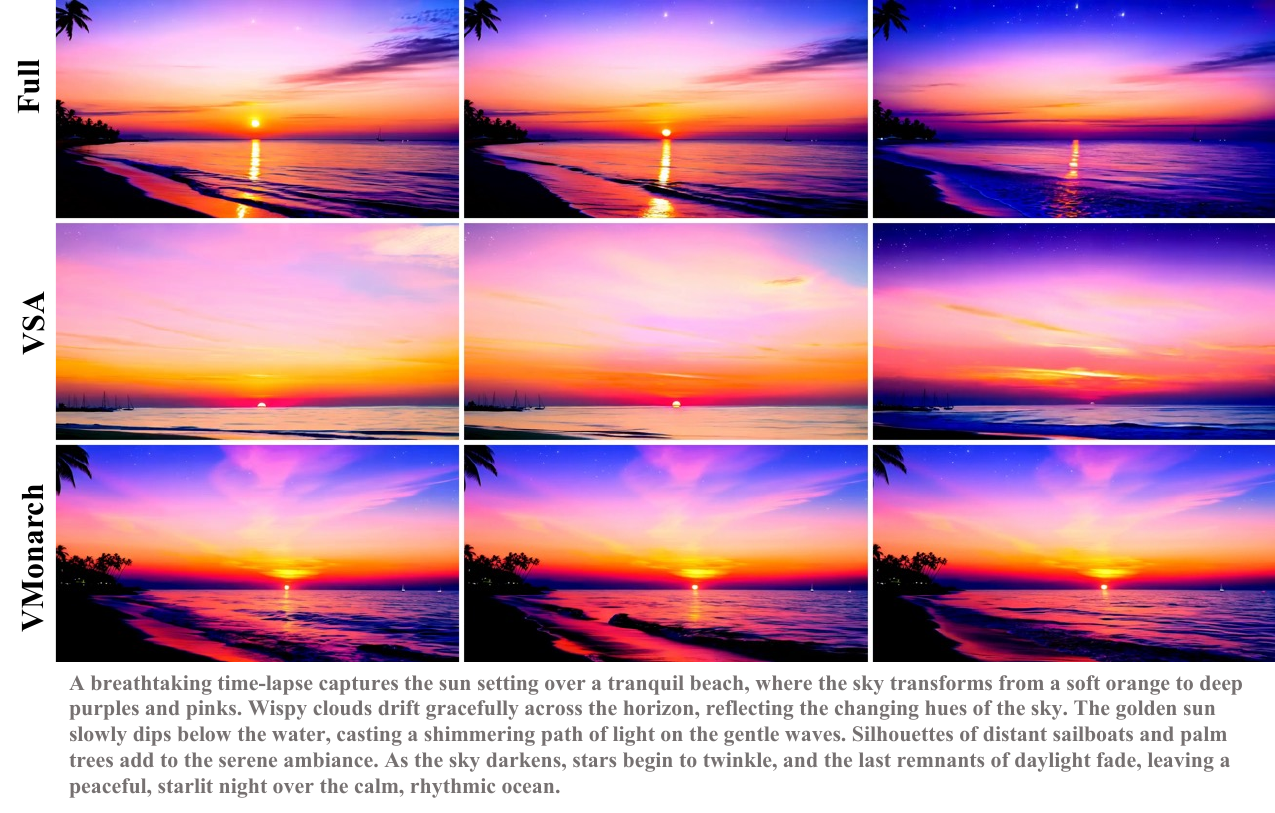}
    \end{subfigure}
    \hfill
    \begin{subfigure}{0.45\linewidth}
        \includegraphics[width=\linewidth]{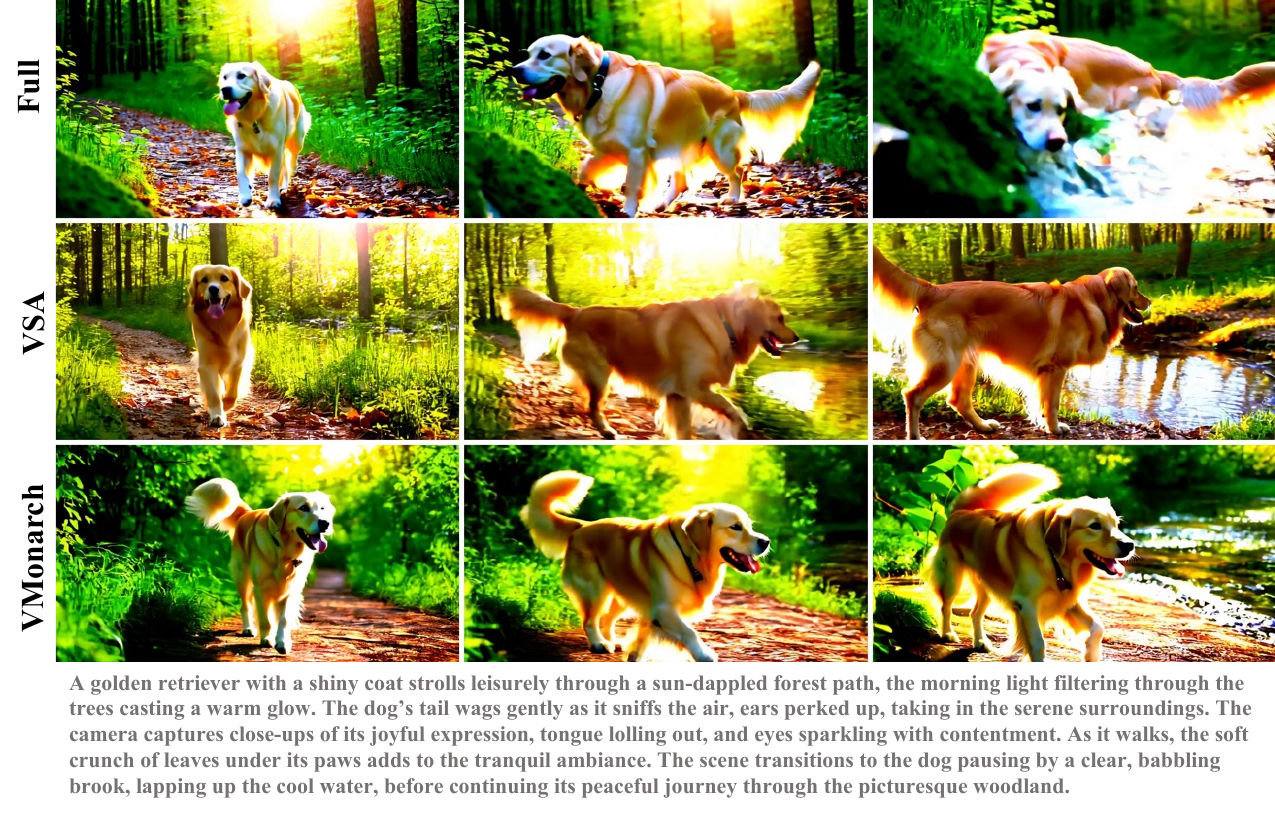}
    \end{subfigure}
    \hfill
    \vspace{1mm}

    \begin{subfigure}{0.45\linewidth}
        \includegraphics[width=\linewidth]{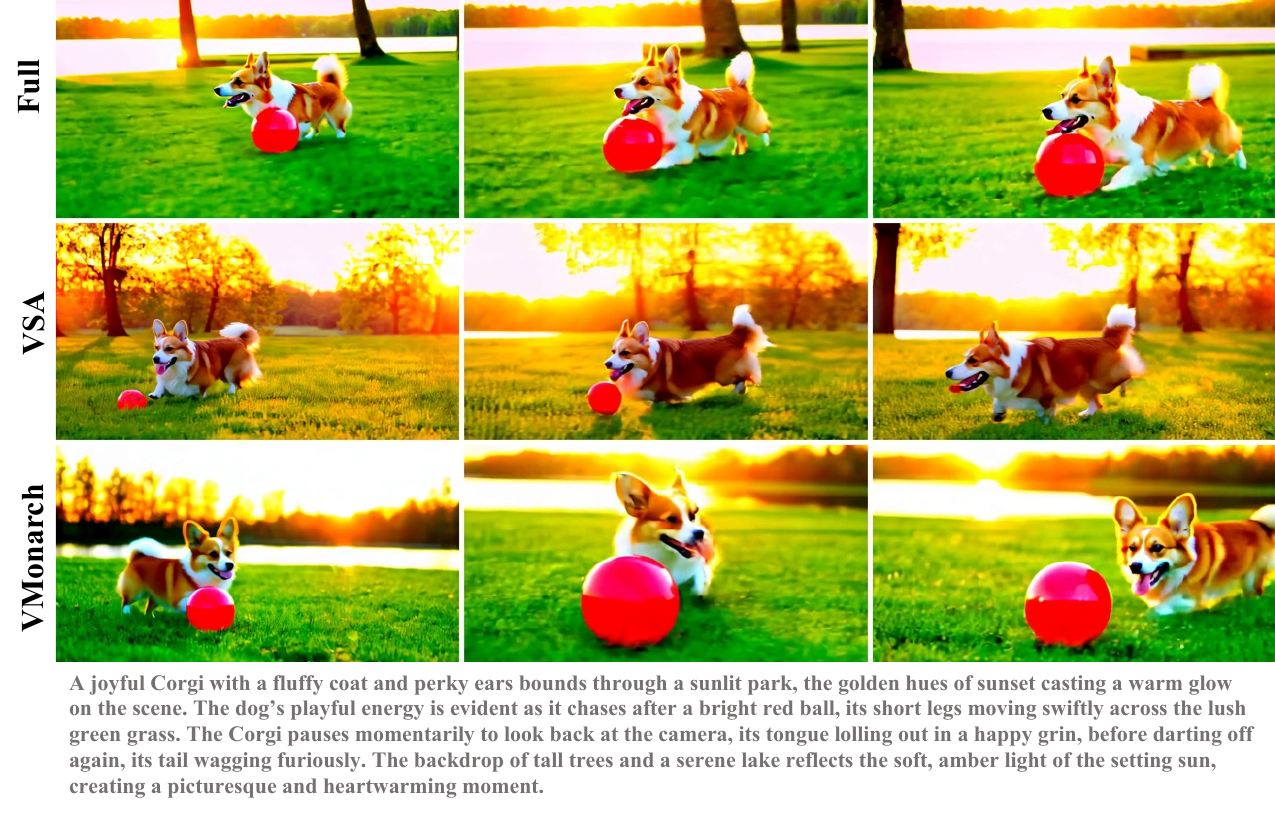}
    \end{subfigure}
    \hfill
    \begin{subfigure}{0.45\linewidth}
        \includegraphics[width=\linewidth]{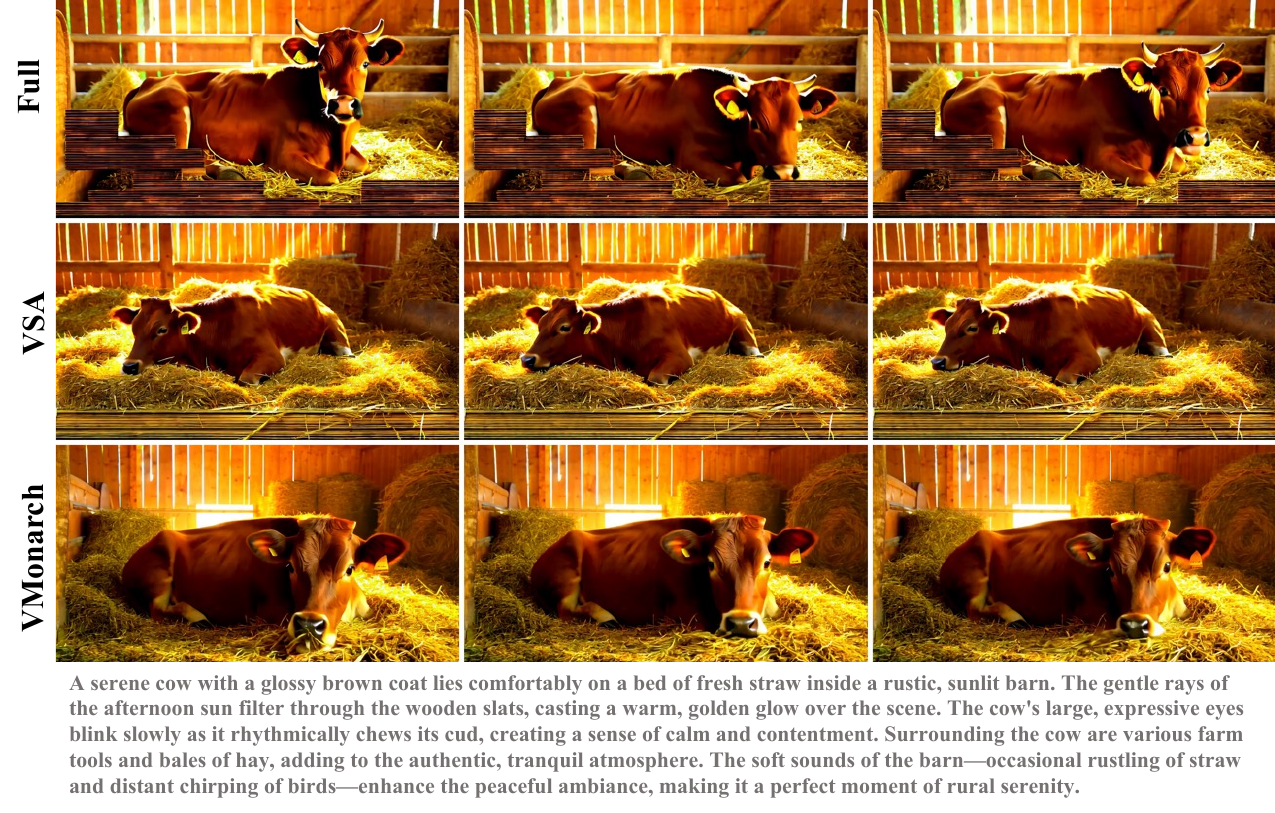}
    \end{subfigure}
    \hfill
    \vspace{1mm}
    
    % Row 2
    \begin{subfigure}{0.45\linewidth}
        \includegraphics[width=\linewidth]{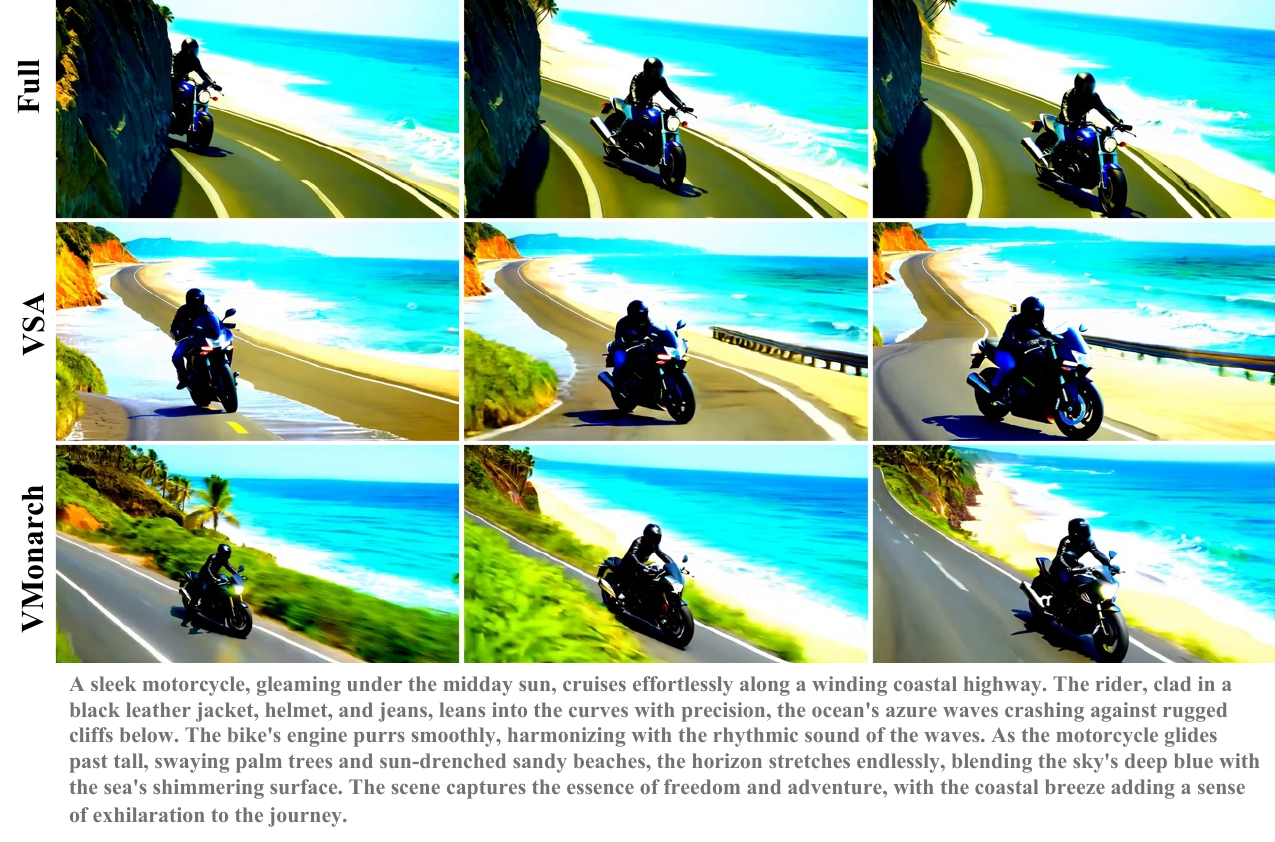}
    \end{subfigure}
    \hfill
    \begin{subfigure}{0.45\linewidth}
        \includegraphics[width=\linewidth]{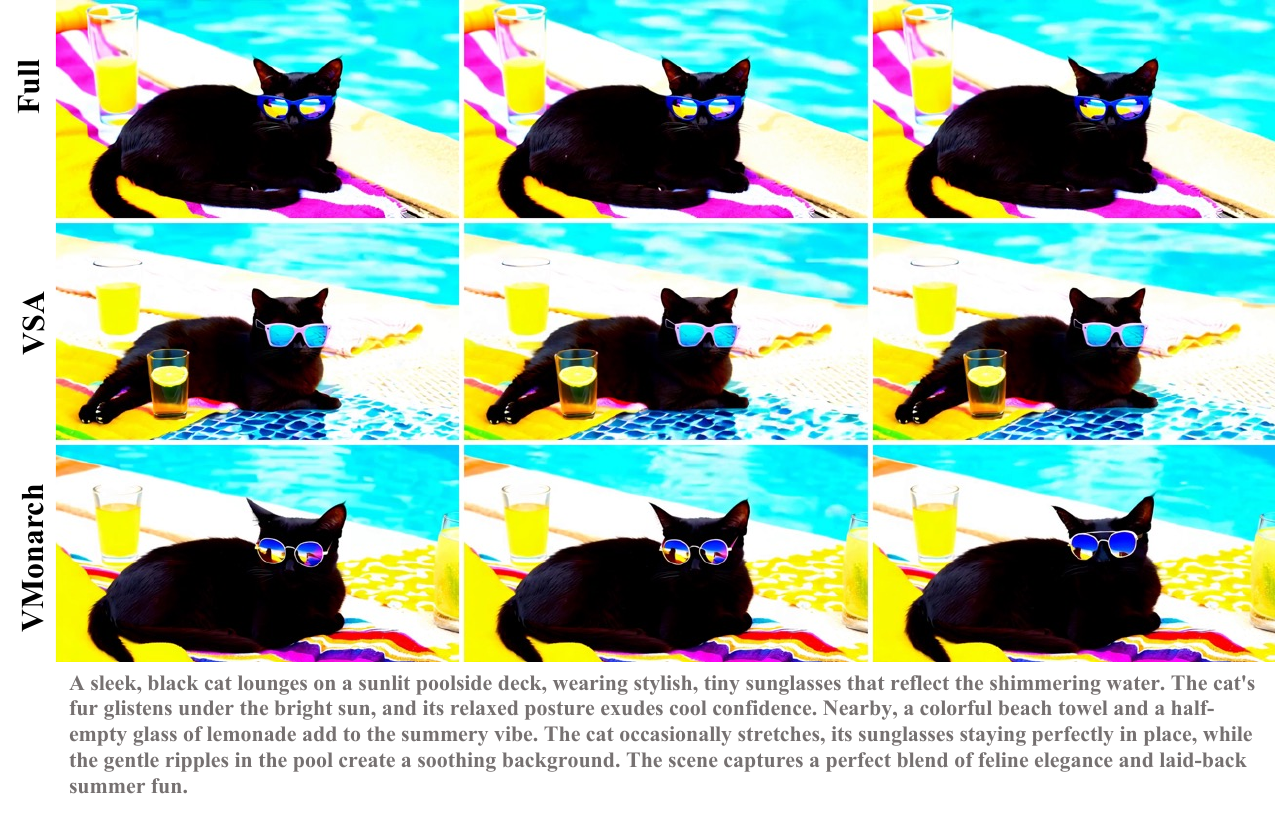}
    \end{subfigure}
    \hfill
    \vspace{1mm}

    \begin{subfigure}{0.45\linewidth}
        \includegraphics[width=\linewidth]{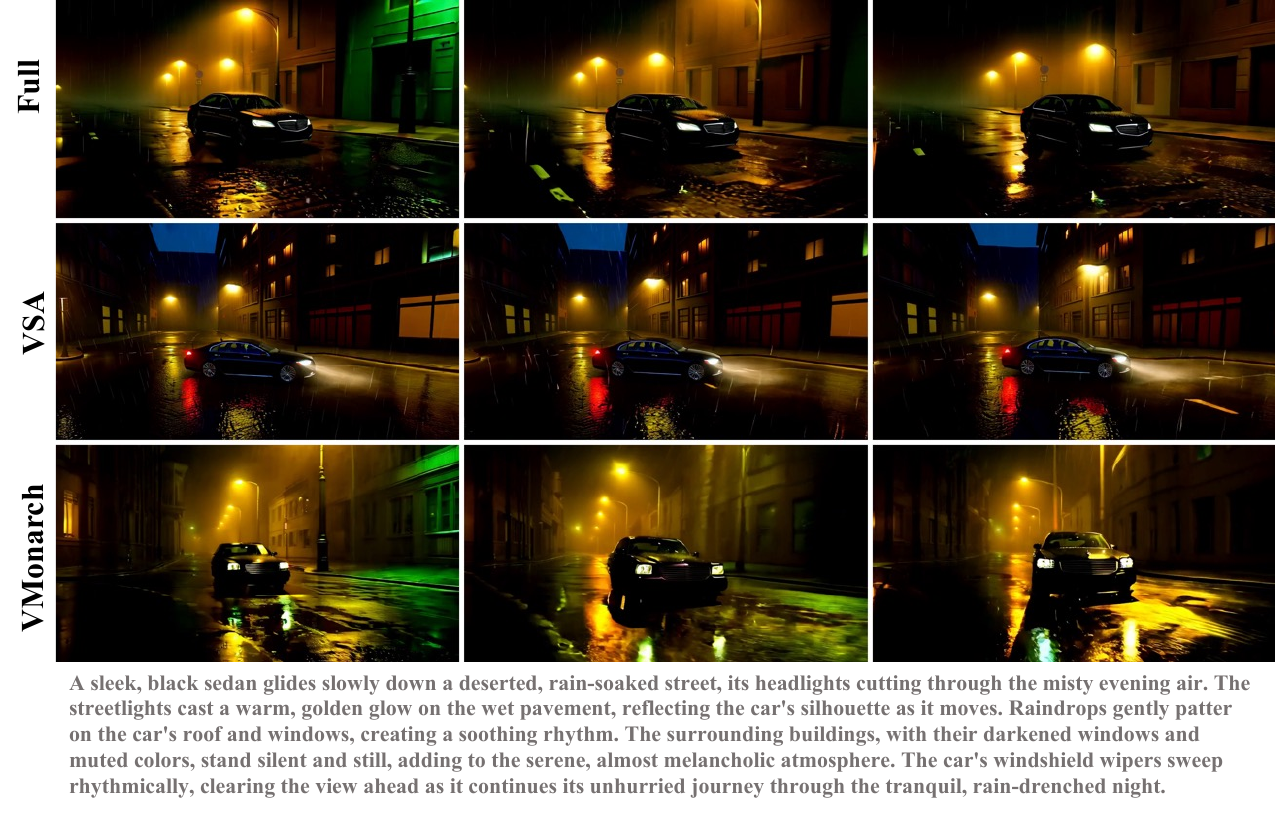}
    \end{subfigure}
    \hfill
    \begin{subfigure}{0.45\linewidth}
        \includegraphics[width=\linewidth]{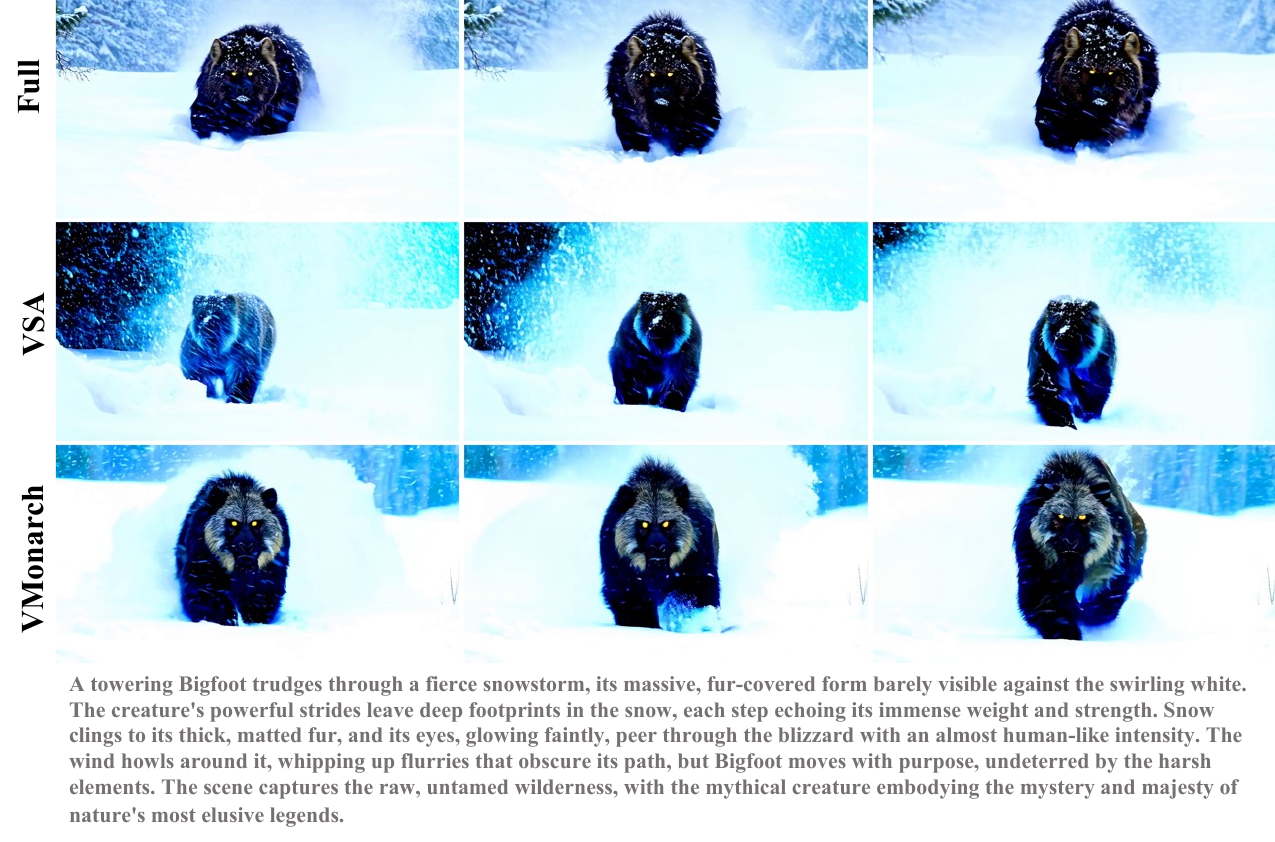}
    \end{subfigure}
    \hfill
    \vspace{1mm}

    \caption{More visual results comparing Full Attention, VSA, and VMonarch on various prompts.}
    \label{fig:more_visual_results}
\end{figure*}

We provide more visual comparisons in~\cref{fig:more_visual_results}, comparing Full Attention, VSA, and our VMonarch on various prompts. The results demonstrate that VMonarch consistently produces higher-quality videos that compare favorably to Full Attention, while VSA sometimes exhibits semantic inconsistencies.

\section{Discussion on Monarch Attention}
Monarch Attention~\cite{yaras2025monarchattention} is a general form of normal self-attention with Monarch matrix~\cite{dao2022monarch} representation, which is suitable for sparse distribution modeling. 
As shown in ~\cref{eq:R}, ~\cref{eq:ar}, ~\cref{eq:L}, ~\cref{eq:al} and ~\cref{eq:MonarchAttention_with_mbt}, Monarch Attention has many parameters and can derive many variants that deserve to be discussed.

\textbf{Monarch Approximation Error Analysis}.
We formulate the Monarch approximation error as  $D_{KL}(S || M) \approx D_{KL}(S || M^*) + D_{KL}(M^* || M)$.
The \underline{structural error} $D_{KL}(S || M^*)$ represents the intrinsic gap between full attention $S$ and the optimal Monarch factorization $M^*$, which is minimized when $S$ aligns with block-diagonal structures; VMonarch effectively reduces this via Spatio-Temporal factorization that matches video priors.
The \underline{iteration error} $D_{KL}(M^* || M)$ stems from the iterative solver, and we find that it converges after 2 iterations.

\textbf{Understanding iteration terms}
$\bm {c_R}$ and $\bm{c_L}$ can be interpreted as temperature adjustment factors for the softmax function based on entropy statistics within each monarch factor $L$ and $R$. They are the intermediate terms under the alternating maximization optimization process. A better optimization algorithm or a suitable optimization object target may bring simpler and more efficient intermediate terms.

\textbf{Correlation with full attention }
As $N=m\times b$, if $b=1$ or $m=1$, MonarchAttention reduces to standard attention.
From this point of view, full attention can be viewed as a special case of Monarch attention, where only a single monarch factor is applied.

\textbf{Correlation with spatial-temporal attention}
When no iteration is performed, Monarch Attention degenerates to Spatial/Temporal Attention.
When only $R$ updating~\cref{eq:R} is applied and $\mathbf{c}_{\mathbf{R}}$ is set to the all-ones matrix and $L$ is set to the identity matrix,  it is equal to the temporal-attention.

Although Monarch Attention~\cite{yaras2025monarchattention} provides an iterative approximation algorithm to reduce $D_{KL}(M^* || M)$, its inefficient entropy calculation hinders scaling to long sequences.
Our proposed \underline{Online-Entropy Flash Attention}
overcomes the OOM bottleneck via algorithm-system co-design, which is critical for the video modality.
Furthermore, we leverage \underline{Spatio-Temporal Factorization} aligned with video priors to directly minimize  $D_{KL}(S || M^*)$, and introduce a general \underline{Recomputation} strategy to help model convergence by losslessly approximating specific tokens.

\section{Limitations and Future Work}
VMonarch has several limitations that present opportunities for future research. 
First, we employ a static sparsification strategy with uniform Monarch parameters across all layers and heads, whereas a dynamic strategy that adapts to the distinct distribution priors of different layers could potentially improve performance. Second, although VMonarch achieves a great theoretical sparsity, it still relies on dense attention within each Monarch factor, and sparsifying this internal computation could further boost efficiency. Third, a more balanced Monarch factorization (e.g., via spatial downsampling) could further enhance efficiency.

\end{document}